%% file: 0_main.tex
\documentclass[10pt,twocolumn,letterpaper]{article}

\usepackage[pagenumbers]{cvpr} 
\usepackage{graphicx}
\usepackage{amsmath}
\usepackage{amssymb}
\usepackage{booktabs}
\usepackage[symbol]{footmisc}
\usepackage[pagebackref,breaklinks,colorlinks]{hyperref}

\usepackage[capitalize]{cleveref}
\crefname{section}{Sec.}{Secs.}
\Crefname{section}{Section}{Sections}
\Crefname{table}{Table}{Tables}
\crefname{table}{Tab.}{Tabs.}

\input{0_macros}

\def\cvprPaperID{4828} 
\def\confName{CVPR}
\def\confYear{2024}

\hypersetup{
    urlcolor=blue 
}
\begin{document}

\title{\includegraphics[width=0.9em]{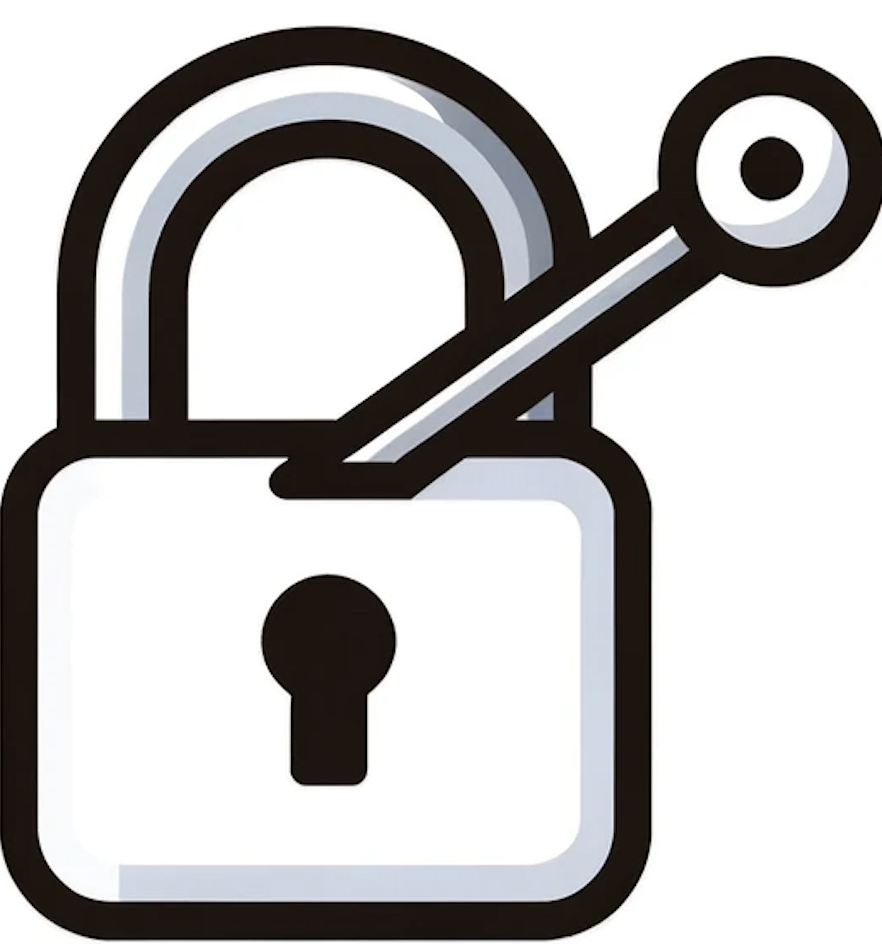 } PIN: Positional Insert Unlocks Object Localisation Abilities in VLMs}

\author{Michael Dorkenwald
 \quad Nimrod Barazani \quad Cees G. M. Snoek\thanks{Equal last author. 
 }
 \quad Yuki M. Asano\footnote[1]{} \\
University of Amsterdam 
\\ 
\href{https://quva-lab.github.io/PIN/}{https://quva-lab.github.io/PIN/}
}

\maketitle
\input{1_abstract}

\input{2_intro}

\input{3_relatedwork}

\input{3_study}

\input{4_method}

\input{5_experiments}

\input{6_conclusions}


{\small
\bibliographystyle{ieee_fullname}
\bibliography{ref}
}

\clearpage
{\noindent \Huge \textbf{Supplemental}}

\appendix
\input{7_supp}

\end{document}

%% file: 0_macros.tex
\usepackage{graphicx}
\usepackage{amssymb}
\usepackage{amsmath}
\usepackage{booktabs}

\usepackage{times}
\usepackage{epsfig}
\usepackage{tabu}
\usepackage{multirow}
\usepackage{amsfonts}       
\usepackage{xpunctuate}
\usepackage{pifont}
\usepackage{color,colortbl}
\usepackage[rgb,dvipsnames]{xcolor}
\usepackage{tabulary,overpic}
\usepackage{booktabs}       
\usepackage[normalem]{ulem}

\usepackage{scalerel}
\usepackage[accsupp]{axessibility}

\usepackage{enumitem}
\usepackage{xspace}
\usepackage{booktabs}
\usepackage{colortbl}
\usepackage{multirow}
\usepackage{makecell}
\usepackage{lipsum} 
\usepackage{gensymb} 

\definecolor{fyxcolor}{RGB}{0,128,255}
\definecolor{demphcolor}{RGB}{100,100,100}
\definecolor{citecolor}{RGB}{0,0,192} 
\definecolor{GrayBG}{gray}{0.95}

\newcommand{\app}{\raise.17ex\hbox{$\scriptstyle\sim$}}
\newlength\savewidth

\newcommand{\fref}[1]{Fig.~\ref{#1}}

\newrobustcmd{\B}{\bfseries}

\usepackage[labelsep=period]{caption}
\renewcommand{\paragraph}[1]{\vspace{0.2em}\noindent \textbf{#1 \hspace{0.2em}}}
\definecolor{lightcyan}{rgb}{0.88, 1.0, 1.0}

\definecolor{MyDarkRed}{rgb}{0.66, 0.16, 0.16}
\definecolor{MyDarkBlue}{rgb}{0.16, 0.16, 0.66}



%% file: 1_abstract.tex
\begin{abstract}
Vision-Language Models (VLMs), such as Flamingo and GPT-4V, have shown immense potential by integrating large language models with vision systems. Nevertheless, these models face challenges in the fundamental computer vision task of object localisation, due to their training on multimodal data containing mostly captions without explicit spatial grounding. While it is possible to construct custom, supervised training pipelines with bounding box annotations that integrate with VLMs, these result in specialized and hard-to-scale models. In this paper, we aim to explore the limits of caption-based VLMs and instead propose to tackle the challenge in a simpler manner by i) keeping the weights of a caption-based VLM frozen and ii) not using any supervised detection data. 
To this end, we introduce an input-agnostic Positional Insert (PIN), a learnable spatial prompt, containing a minimal set of parameters that are slid inside the frozen VLM, unlocking object localisation capabilities. Our PIN module is trained with a simple next-token prediction task on synthetic data without requiring the introduction of new output heads. Our experiments demonstrate strong zero-shot localisation performances on a variety of images, including Pascal VOC, COCO, LVIS, and diverse images like paintings or cartoons. 
\end{abstract}

\vspace{-0.5cm}

%% file: 2_intro.tex
\section{Introduction}
\label{sec:intro}
Vision-Language Models (VLMs) have shown remarkable results across diverse tasks, propelled by the advancements in Large Language Models (LLMs)~\cite{llama_touvron, brown_gpt, chowdhery2022palm}. Early works~\cite{radford2021learning, jia_scale, lxmert_tan, albef_lavis, coca_yu} used extensive image-caption data for end-to-end training, a trend later evolved by works like~\cite{blip1, blip2, alayrac2022flamingo, koh2023grounding, lit_clip, pali}, which efficiently integrated pretrained vision and language models through fusion networks to further enhance cross-modal understanding.
Flamingo~\cite{alayrac2022flamingo} demonstrates impressive multimodal in-context learning abilities. However, like many caption-based VLMs, it faces challenges in object localisation, a consequence of its training on web data. 

Equipping VLMs with precise object localisation abilities is important for tasks like autonomous driving~\cite{wayve_lingo1,wen2023road, driving_vlm_gpt4}, assistive technology~\cite{yang2022seeway}, and robotics~\cite{palme_robotics, vlm_robotics, robotic_control_vlm}. Despite their proficiency in integrating visual-textual data, current image-caption training hinders accurate spatial understanding. Therefore, enhancing spatial comprehension in VLMs is key to enabling more nuanced and context-aware interactions.

One recent stream of research~\cite{vision_llm, unitab, wang2023cogvlm, ofa, mini_gpt_v2, mplug_owl, unified_IO, glipv2} focuses on developing unified \textit{expert} Vision Language Models (VLMs) capable of performing a variety of tasks, including localisation, with a universal architecture. Although these models show impressive results across different tasks, their success largely depends on the availability of extensive task-specific, supervised data~\cite{vision_llm, glip, wang2023cogvlm, mini_gpt_v2}. Furthermore, \cite{unified_IO, mini_gpt_v2, vision_llm, wang2023cogvlm, ofa, kosmos2} require a large amount of compute for training.
The setting we tackle in this paper is different. 
Our goal is to efficiently enable the localisation capabilities of VLMs while keeping their parameters untouched and \textit{without} the need for localisation supervised datasets.

\begin{figure}[tb] \centering
    \includegraphics[width=0.48\textwidth]{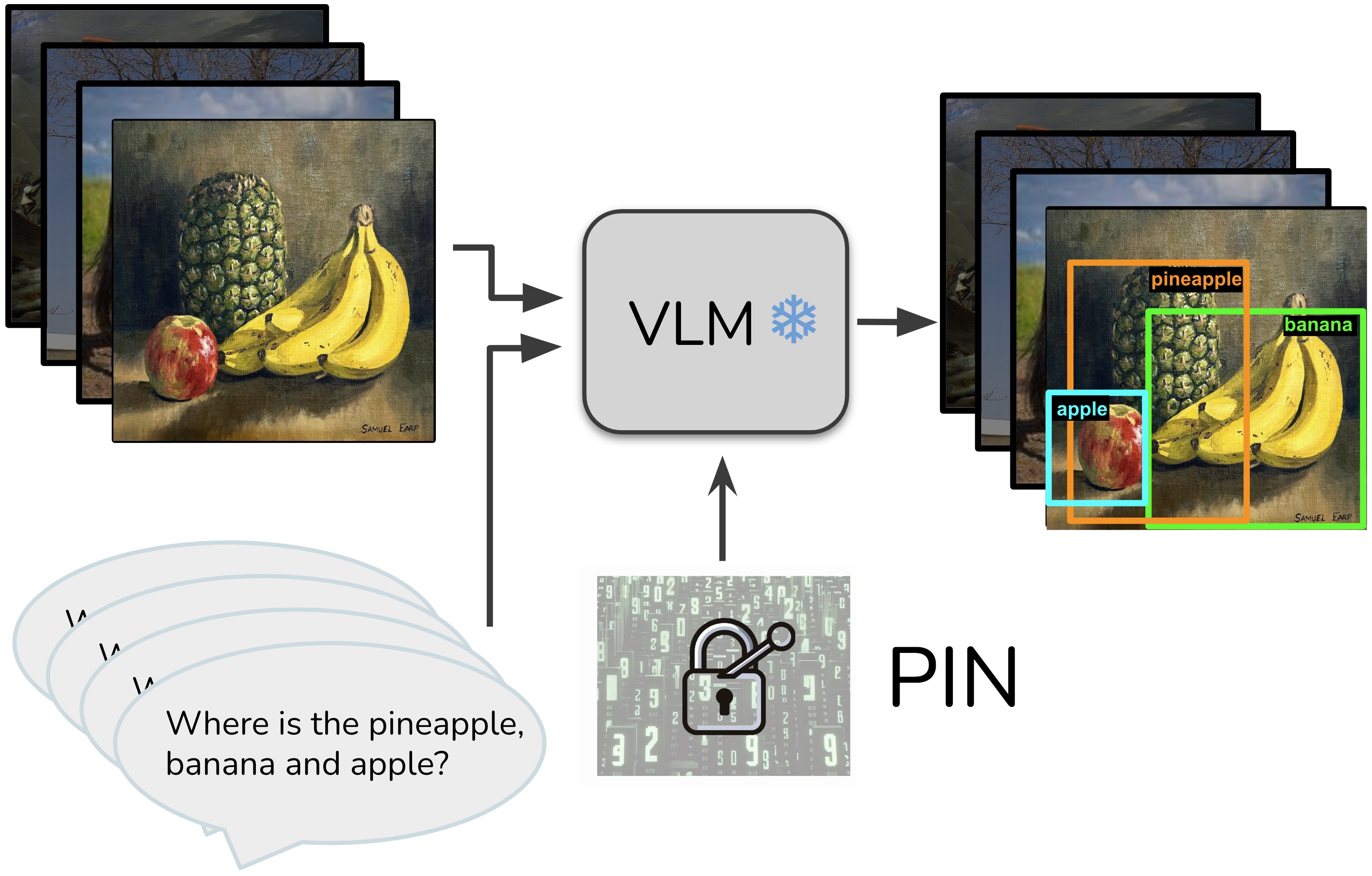}
    \caption{We learn a single Positional Insert (PIN) for unlocking zero-shot object localisation abilities in a frozen Vision Language Model (VLM) without adding any additional heads or requiring supervised datasets. Further output examples shown in Fig.~\ref{fig:ADE} \&~\ref{fig:coco_pvoc}.} \label{fig:teaser}
\vspace{-0.6cm}
\end{figure}
\begin{figure*}[tb] \centering
    \includegraphics[width=\textwidth]{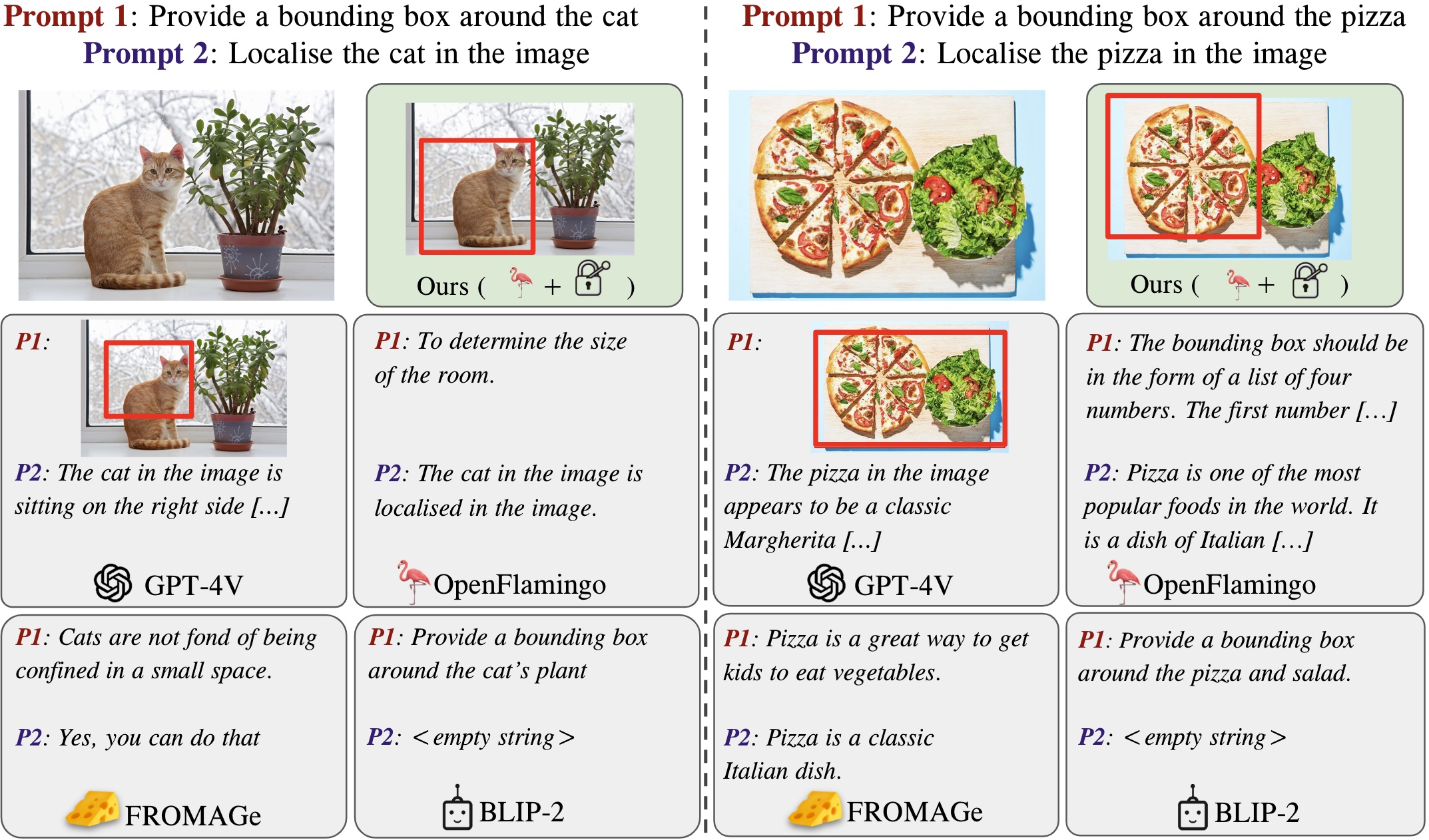}
    \caption{Examples from our analysis on localisation abilities of existing caption-based VLMs. GPT-4V\cite{open_ai-gpt4} is the only model to return bounding boxes and by that roughly localised the object. All other VLMs struggle to easily localise the objects in the image. Further examples and different kinds of prompts are provided in the supplemental Sec.~\ref{sec:extended_study}.
    } \label{fig:study}
\vspace{-0.6cm}
\end{figure*}

Our work aims to unlock the localisation abilities of caption-based VLMs by integrating spatial understanding into their existing zero-shot capabilities. We introduce a Positional Insert (PIN), a learnable spatial prompt designed to infuse spatial awareness into VLMs without altering their pretrained weights. Our learned PIN is simply added to the vision encoder embedding and follows the VLMs forward pass from there, thereby not imposing any computational overhead. 
To train our PIN module effectively and without supervised data, we create a synthetic dataset composed of synthesized object renderings superimposed on background images, providing precise ground truth locations. We assess our approach on COCO~\cite{coco}, PVOC~\cite{pvoc}, LVIS~\cite{lvis}, and RefCOCO~\cite{refcoco}. Our findings reveal a significant enhancement in VLMs’ object localisation abilities. Our contributions can be summarized as follows:
\begin{itemize}
    \item We provide an analysis of the abilities of caption-based VLMs for object localisation.
    \item We propose PIN, a spatial prompt, to unlock the localisation abilities in caption-based VLMs.
    \item We demonstrate on the OpenFlamingo~\cite{openflamingo_awadalla} and BLIP-2~\cite{blip2} VLMs the ability to successfully localise objects on COCO, PVOC, LVIS, and other data.
\end{itemize}

%% file: 3_relatedwork.tex
\begin{figure*}[tb] \centering

    \includegraphics[width=\textwidth]{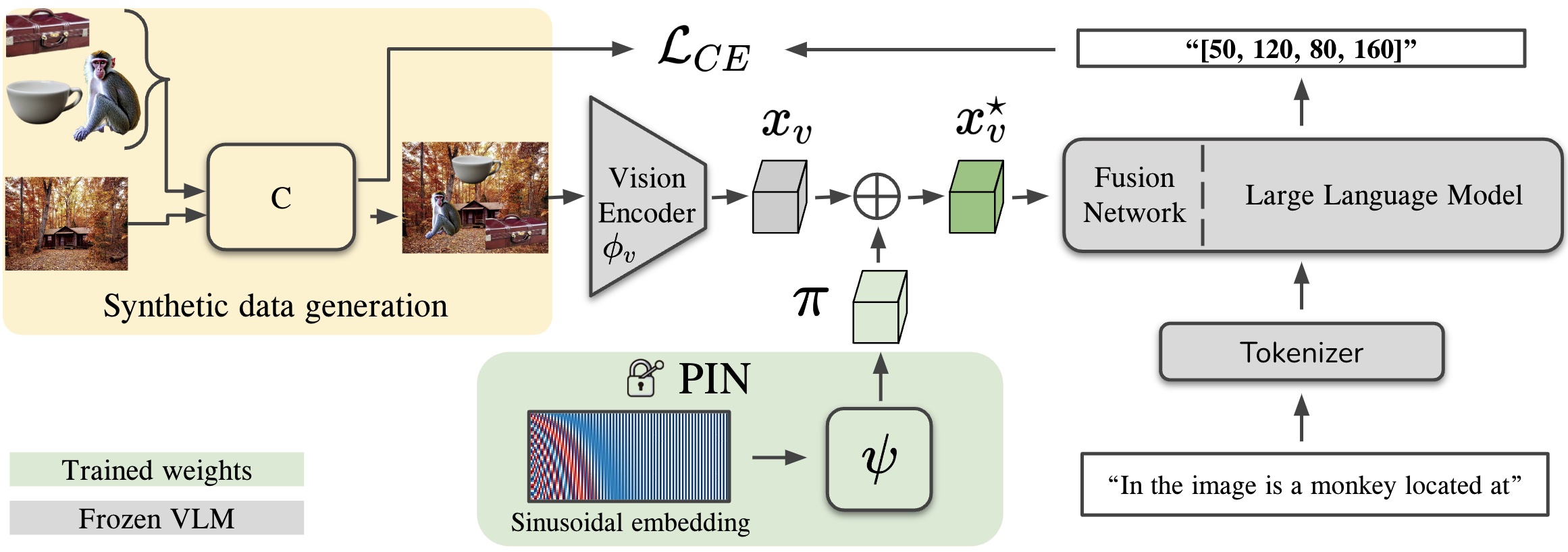}
    \caption{Schematic overview of our method. We generate synthetic training data by overlaying objects on background images using our composition function $C$. These images are then encoded, and our lightweight learnable spatial prompt vector \( \pi \) from the PIN module is added to their vision encodings \(x_v\). Using the VLM's standard forward pass, a location text response is generated based on the input object name and the enhanced visual feature \(x^{\star}_v\). The PIN module is optimized with cross-entropy by comparing this generated text against the known object locations from the composition function $C$. }\label{fig:method}
\vspace{-0.6cm}
\end{figure*}

\vspace{-0.2cm}
\section{Related Work}
\label{sec:related_works}

\paragraph{Caption-based Vision-Language Models.}
Large Language Models (LLMs)~\cite{brown_gpt, open_ai-gpt4, llama_touvron, chowdhery2022palm} have not only been transformative for the field of natural language processing but have also significantly propelled the development of multimodal models. Initial works for Vision Language Models (VLMs)~\cite{alayrac2022flamingo, pali, git_wang, SIM_VLM, blip2, cm3, blip1} concentrated on extensive image-text pretraining. These models typically undergo pretraining with vast collections of interleaved image-text data~\cite{c4_data, schuhmann2022laion}. Flamingo was a pioneer in merging a pretrained CLIP~\cite{radford2021learning} image encoder with a pretrained LLM through a perceiver and gated cross-attention blocks, demonstrating strong multimodal in-context learning abilities. Given the image-text pretraining data containing descriptive captions for images, we categorize these VLMs as caption-based. This kind of pretraining naturally limits the spatial comprehension and expression abilities of those VLMs. In this paper, we present a new, simple, and efficient way designed to enable object localisation capabilities within these models.

\paragraph{Expert-based Vision-Language Models.} 
Universal frameworks \cite{pix2seq, unified_IO, glip, unitab, gato} have been introduced to unify architectures and training tasks by treating it as a language modeling problem conditioned on e.g. observed pixel inputs. Recent works~\cite{instr_blip, mini_gpt4, otter_li, koh2023grounding, wang2023cogvlm} applied this to multimodal instruction-tuned data, promoting more intuitive human-model interactions for VLMs. The resulting \textit {unified expert} VLMs are capable of handling diverse tasks. Many others \cite{vision_llm, unitab, ofa, mini_gpt_v2, mplug_owl, unified_IO, wang2023cogvlm, glipv2, kosmos2} additionally target visual grounding tasks like localisation. Yet, those VLMs rely on large annotated localisation datasets \cite{objects365, coco, visual_genome, refcoco}. In addition, many of those works\cite{kosmos2, glipv2, unified_IO, vision_llm, unitab} require substantial amounts of compute to leverage this data. 
While these models exhibit impressive performance across various tasks, hence the name experts, their success hinges on large quantities of task-specific, supervised data and computational resources.
Our work diverges from this path, seeking to unlock the object localisation capabilities of caption-based VLMs without relying on manually annotated datasets. We propose a more flexible and efficient strategy, exploring how far we can go without supervised data.

\paragraph{Visual Prompt Learning.}
Prompt Learning is a method originated from NLP~\cite{prefix_Li, PL_NLP_lester, PL_NLP_Liu} where prompts are viewed as continuous, task-specific vectors optimized during finetuning. This technique matches the performance of full finetuning but requires 1000 times fewer parameters, enhancing efficiency and reducing resource usage. Beginning works focused on adapting those methods to VLMs by adding learnable tokens to the language model~\cite{zhou2022cocoop, PL_gradient_Zhu, zhou2022coop, da_pl}. Subsequent works~\cite{VPT_belongie, PL_continual_Wang, Video_PL_Ju, PL_distribution_LU, VP_pixel_Isola} extended them to the vision model and recently to both the vision and language branch~\cite{maple_MLPL}. However, these works have been applied to encoder-only models, such as CLIP \cite{radford2021learning}, leaving their adaption to VLMs with a decoder unexplored. Motivated by these methods, we introduce a positional prompt for specifically targeting localisation in generative VLMs.


%% file: 3_study.tex
\section{Localisation by Caption-based VLMs}
\label{sec:study}
Before discussing our proposed method, we first assess the object localisation capabilities of caption-based VLMs by analysing their textual responses given various prompts. We examine models such as GPT-4V~\cite{open_ai-gpt4}, BLIP-2~\cite{blip2}, Flamingo~\cite{openflamingo_awadalla, alayrac2022flamingo}, and Fromage~\cite{koh2023grounding}. For that, we use prompts aimed at generating a bounding box response from these VLMs. Note that due to the undisclosed training data for \mbox{GPT-4V}~\cite{open_ai-gpt4}, we cannot rule out its exposure to supervised object localisation training. We compare this against the publicly available 9B version of OpenFlamingo~\cite{openflamingo_awadalla} and the 7B version of BLIP-2~\cite{blip2}. An overview of the results and prompts can be found in \fref{fig:study}. 
We find that among the evaluated VLMs, only GPT-4V~\cite{open_ai-gpt4} successfully returns bounding boxes that roughly localise the intended object. Other VLMs~\cite{koh2023grounding, openflamingo_awadalla, blip2} are unable to provide any location information even in text form and instead are ``chatty'' (FROMAGe, OpenFlamingo) or return the input or provide no output (BLIP-2). 
In Sec.~\ref{sub:Quantitative_Results}, we quantitatively evaluate the in-context learning abilities for localisation of the OpenFlamingo model.  
In the supplementary material Sec.~\ref{sec:extended_study}, we broaden our study by examining a wider variety of prompts, specifically including those that do not require generating a bounding box, and by analyzing a larger number of samples. Yet, the conclusion remains the same as with the exemplary results in Fig.~\ref{fig:study} that caption-based VLMs are unable to localise objects in a given image via textual responses. 

%% file: 4_method.tex
\section{Method}%
\label{sec:method}


We tackle the shortcomings of caption-based Vision-Language Models (VLMs) in their ability to localise objects within images. To this end, we introduce a simple yet effective Positional Insert (PIN), designed to enhance the VLMs' object localisation capabilities without altering their existing parameters. An overview of our approach can be found in Fig.~\ref{fig:method}.

\paragraph{Preliminary.}
Vision-Language Models (VLMs) accept inputs composed of visual data such as images \( I \) alongside a textual input \( T \). The visual component \( I \) is processed by a vision encoder \( \phi_V \) producing a feature vector \( x_v \in \mathbb{R}^{N_p \times D_v} \), where \( N_p \) denotes the number of patches and \( D_v \) the channel dimension. Similarly, the textual information \( T \) is tokenized, yielding textual embeddings \( x_{\text{t}} \in \mathbb{R}^{M \times D_t} \), with \( M \) representing the amount of textual tokens and \( D_V \) the vocabulary size. The visual features \( x_v\) go through a fusion network $F$ before being processed with the textual features \( x_t \) to produce a response text \( t_r {=} \text{LLM}(F(x_v), x_t) \) by the Large Language Model.

\subsection{PIN: Positional Insert}
\label{subsec:PIN}
The Positional Insert is a learnable input-agnostic spatial feature vector and is inserted directly after the vision encoder \( \phi_V \). To instill spatial awareness into our PIN, we start with fixed positional embeddings of dimension \( d \) employing sinusoidal functions~\cite{vaswani2017attention}
\begin{align}
S[i, 2k] &= \sin\left(\frac{\text{position}}{10000^{2k/d_{\text{model}}}}\right), \\
S[i, 2k+1] &= \cos\left(\frac{\text{position}}{10000^{2k/d_{\text{model}}}}\right),
\end{align}
where \( i \) denotes the index of the position and \( k \) represents the index within the dimension of the embedding, with \( d_{\text{model}} \) as the dimensionality of the embedding space. The range for \( k \) extends from 1 to \( d_{\text{model}} \). Each of the spatial sinusoidal vectors is further refined by a learnable, shallow feed-forward neural network $\psi$ parametrized by \( \theta \), resulting in our PIN \(\pi {=} \psi(S) \) with the output dimension matching the ones from the vision encoder \( \pi \in \mathbb{R}^{M \times D_t} \). This learned embedding is then added to the output from the vision encoder \( x_v \), resulting in the enriched visual feature representation
\begin{align}
    x_v^\star = x_v + \pi.
\end{align}

\begin{figure}[t] 
    \includegraphics[width=\columnwidth]{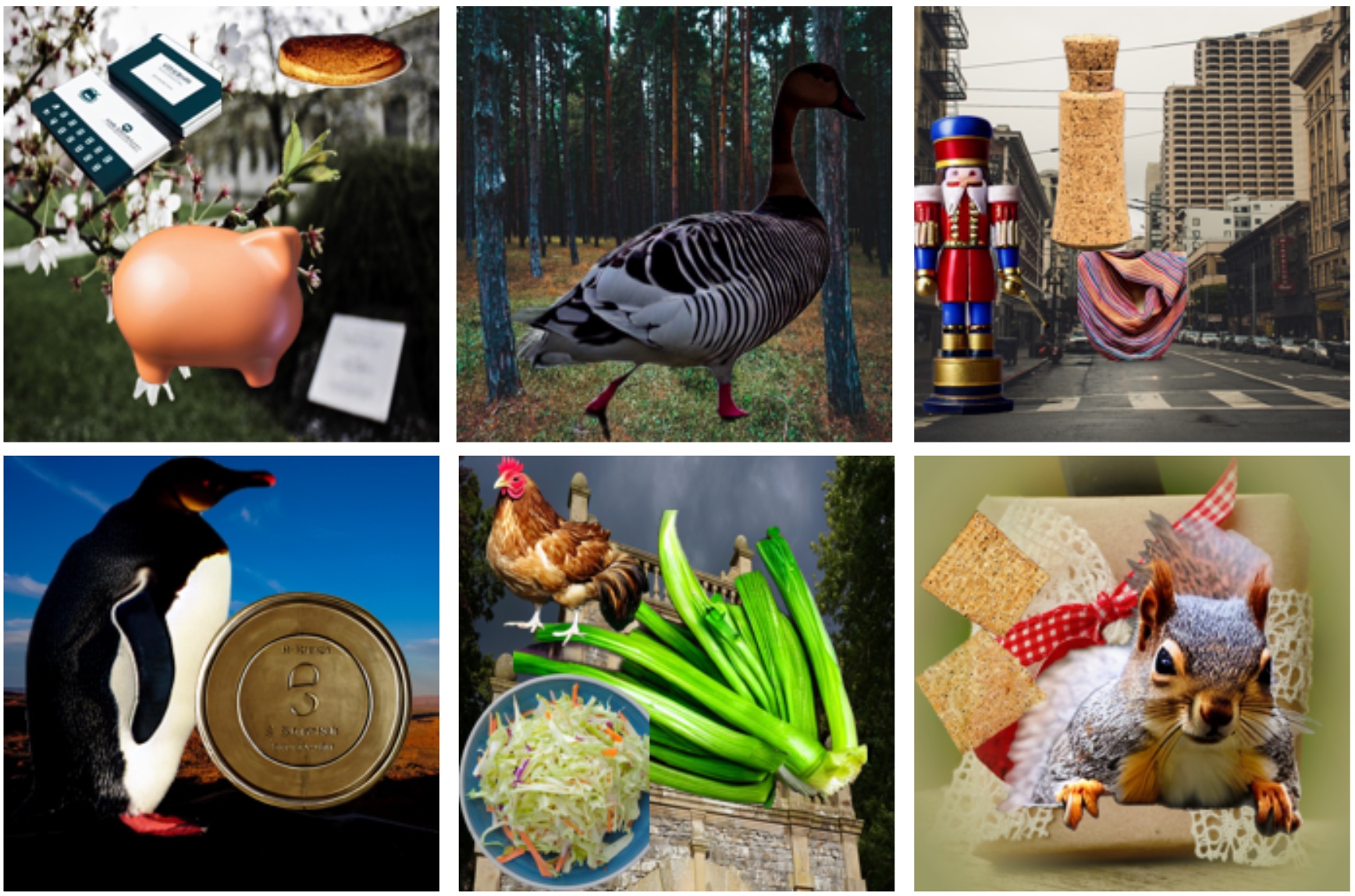}
    \caption{Sample images from our synthetic data generation.} \label{fig:syntehtic_zero_shot}
\vspace{-0.3cm}
\end{figure}

\paragraph{Training Objective.} 
The PIN module's parameters \( \theta \) of \( \psi \) are optimized via the text output produced by the large language model. This process requires no additional heads or projection layers, thus maintaining the model's simplicity and native natural language output. The model is trained with an input sequence consisting of a textual prompt \(t_p \in T\) such as `In the image is a \(<\)\textit{obj}\(>\) located at' and is tasked to complete the sequence with the bounding box coordinates. For a given object name \(<\)\textit{obj}\(>\), present within the image, the model predicts a sequence of bounding box coordinates in the template of \(t_r \in T\) like \([x_{\text{min}}, y_{\text{min}}, x_{\text{max}}, y_{\text{max}}]\) conditioned on the image features and the initial textual prompt. We employ a negative log-likelihood loss for the predicted tokens
\begin{equation}
    \mathcal{L}_{CE}(\theta) = -\sum_{t=1}^{T} \log p_\theta \left( y_t | y_{<t}, x_v^\star \right),
\end{equation}
where \( y_t \) corresponds to the target token at position \( t \) in the text, \( T \) is the total number of tokens to be predicted and $x_v^\star$ is the positional enhanced feature vector. Here \( p_{\theta} \) is the probability assigned by the model to the correct token at position \( t \), conditioned on the previous tokens \( y_{<t} \), the visual features, and the textual prompt. This learning objective enables the easy adaption of pretrained VLMs for localisation without the dependency on specialized components like region proposal networks.

\input{tables/main}

\subsection{Synthetic Data Generation}
We do not rely on manually labeled data to unlock the positional information in the VLM. Instead, we generate our own synthetic data following \cite{xpaste, initial_xpaste} by utilizing Stable Diffusion~\cite{rombach_stable_diffusion} to synthesize objects from the LVIS~\cite{lvis} category list. 
The CLIP~\cite{clip_icml21} module is used to sort out implausible images by removing those with a low CLIP~\cite{clip_icml21} score, a matching score between the input image I and the textual information T. Note, since the vision encoder's weights remain unchanged, it is unlikely to overfit to any pasting artifacts. The composition function \( C \) overlays objects on randomly picked locations while considering the following constraints: the aspect ratio \( r \) of objects, minimal \( s_{\text{min}} \) and maximal \( s_{\text{max}} \) pasting sizes, the number of objects \( a_{\text{max}} \), and the maximal overlap \( o_{\text{max}} \) \wrt already inserted objects. Given a background image \( I_b \in I \), the composition function yields
\begin{align}
    (t_p, I_p) = C(I_b, r, a_{\text{max}}, s_{\text{min}}, s_{\text{max}}, o_{\text{max}}), 
\end{align}
with a generated image \(I_p\in I\) and the text \( t_p \in T \) containing the object location for a randomly selected object by \( C \). This process creates a self-generated supervision signal that is subsequently exploited in the training of PIN. Typical sample images can be found in Fig. \ref{fig:syntehtic_zero_shot}.

%% file: tables/main.tex
\begin{table*}[t]
\centering
\resizebox{\textwidth}{!}{%
\begin{tabular}{ll ccc | ccc | ccc }
\toprule
&\multirow{2}{*}{Method} & \multicolumn{3}{c}{$\text{PVOC}_{\leq 3 \text{ Objects}}$ } & \multicolumn{3}{c}{$\text{COCO}_{\leq 3 \text{ Objects}}$ } & \multicolumn{3}{c}{$\text{LVIS}_{\leq 3 \text{ Objects}}$ }\\
&& mIoU & $\text{mIoU}_M$ & $\text{mIoU}_L$ & mIoU & $\text{mIoU}_M$ &  $\text{mIoU}_L$ &  mIoU & $\text{mIoU}_M$ &  $\text{mIoU}_L$  \\
\midrule
&  \textit{Baselines} \\
&  \hspace{2mm} raw  & 0 & 0 & 0 & 0 & 0 & 0 & 0 & 0 & 0 \\
&  \hspace{2mm} random & 0.22\footnotesize{$\textcolor{gray}{\pm 0.04}$} & 0.10\footnotesize{$\textcolor{gray}{\pm 0.02}$} & 0.33\footnotesize{$\textcolor{gray}{\pm 0.06}$}  & 0.12\footnotesize{$\textcolor{gray}{\pm 0.04}$} & 0.07\footnotesize{$\textcolor{gray}{\pm 0.02}$} & 0.22\footnotesize{$\textcolor{gray}{\pm 0.08}$} & 0.07\footnotesize{$\textcolor{gray}{\pm 0.03}$} & 0.06\footnotesize{$\textcolor{gray}{\pm 0.02}$} & 0.18\footnotesize{$\textcolor{gray}{\pm 0.09}$}  \\
\parbox[t]{2mm}{\multirow{5}{*}{\rotatebox[origin=c]{90}{ OpenFlamingo~\cite{openflamingo_awadalla}}}} &  \hspace{2mm} 2 context & 0.19\footnotesize{$\textcolor{gray}{\pm 0.11}$} & 0.08\footnotesize{$\textcolor{gray}{\pm 0.05}$} & 0.30\footnotesize{$\textcolor{gray}{\pm 0.18}$} & 0.10\footnotesize{$\textcolor{gray}{\pm 0.08}$}  & 0.06\footnotesize{$\textcolor{gray}{\pm 0.04}$} & 0.18\footnotesize{$\textcolor{gray}{\pm 0.16}$} & 0.04\footnotesize{$\textcolor{gray}{\pm 0.06}$} & 0.03\footnotesize{$\textcolor{gray}{\pm 0.04}$}& 0.10\footnotesize{$\textcolor{gray}{\pm 0.15}$}\\
&  \hspace{2mm} 5 context & 0.19\footnotesize{$\textcolor{gray}{\pm 0.09}$} & 0.07\footnotesize{$\textcolor{gray}{\pm 0.04}$} & 0.31\footnotesize{$\textcolor{gray}{\pm 0.15}$} & 0.10\footnotesize{$\textcolor{gray}{\pm 0.08}$} & 0.06\footnotesize{$\textcolor{gray}{\pm 0.04}$} & 0.20\footnotesize{$\textcolor{gray}{\pm 0.16}$} & 0.06\footnotesize{$\textcolor{gray}{\pm 0.05}$}  & 0.04\footnotesize{$\textcolor{gray}{\pm 0.03}$} & 0.17\footnotesize{$\textcolor{gray}{\pm 0.13}$} \\
& \hspace{2mm} 10 context & 0.20\footnotesize{$\textcolor{gray}{\pm 0.11}$} & 0.06\footnotesize{$\textcolor{gray}{\pm 0.03}$} & 0.32\footnotesize{$\textcolor{gray}{\pm 0.18}$} & 0.09\footnotesize{$\textcolor{gray}{\pm 0.07}$} & 0.05\footnotesize{$\textcolor{gray}{\pm 0.04}$} & 0.17\footnotesize{$\textcolor{gray}{\pm 0.14}$} & 0.05\footnotesize{$\textcolor{gray}{\pm 0.05}$} & 0.03\footnotesize{$\textcolor{gray}{\pm 0.03}$} & 0.15\footnotesize{$\textcolor{gray}{\pm 0.14}$}  \\
\cmidrule{2-11}
& \textit{PEFT} \\
& \hspace{2mm} CoOp on LLM & 0.28 & 0.11 & 0.43 & 0.22 & 0.10 & 0.39 & 0.13 & 0.07 & 0.40 \\
& \hspace{2mm} VPT on $F$ & 0.34 & 0.16 & 0.51 & 0.26 & 0.15 & 0.47 & 0.19 & 0.14 & 0.48 \\
& \hspace{2mm} VPT on $\phi_V$& 0.42 & 0.21 & 0.61 & 0.33 & 0.22 & 0.57 & 0.23 & 0.19 & 0.56 \\
& \hspace{2mm} LoRA on $\phi_V$  & 0.44 & 0.26 & \B 0.62 &0.33 &0.23 & 0.58  & 0.23 & 0.19 & 0.55 \\
& \hspace{2mm} \includegraphics[width=0.9em]{figures/pin_emoji.png } PIN (ours) &\B 0.45 &\B 0.27 &\B 0.62 & \B 0.35 &\B 0.26 & \B 0.59 & \B 0.26& \B 0.24& \B 0.61 \\
\midrule


& \textit{PEFT} \\
& \hspace{2mm} VPT on $F$& 0.33 & 0.12 & 0.51 & 0.27 & 0.12 & 0.50 & 0.18 & 0.11  & 0.47 \\
\parbox[t]{0mm}{\multirow{-3}{*}{\rotatebox[origin=t]{90}{\,\,\,\,BLIP-2~\cite{blip2}}}}& \hspace{2mm} VPT on $\phi_V$  & 0.32 & 0.12 & 0.50 & 0.26 & 0.11 & 0.48 & 0.17 & 0.10  & 0.46 \\
& \hspace{2mm} \includegraphics[width=0.9em]{figures/pin_emoji.png } PIN (ours)  & \B 0.44  &\B 0.24 & \B 0.63 &\B 0.34&\B 0.22 &\B 0.60 & \B 0.26 & \B 0.23 & \B 0.60\\
\bottomrule

\end{tabular}
}
\caption{Comparison on object localisation on a subset of PVOC~\cite{pvoc}, COCO~\cite{coco} and LVIS~\cite{lvis} with up to 3 objects per image, yielding 3,582, 2,062 and 6,016 test images respectively. PIN improves on the OpenFlamingo in-context and PEFT baselines for both the OpenFlamingo and BLIP-2 VLM.}
\label{tab:method_comparison}
\vspace{-0.5cm}
\end{table*}

%% file: 5_experiments.tex
\section{Experiments}%
\label{sec:Experiments}
We apply our approach to the Flamingo~\cite{alayrac2022flamingo} and BLIP-2~\cite{blip2} VLM. More specifically we use the open-source version OpenFlamingo~\cite{openflamingo_awadalla} for Flamingo. We evaluate the localisation abilities of our approach on a subset of COCO~\cite{coco}, PVOC~\cite{pvoc}, and LVIS~\cite{lvis} with up to 3 objects per image resulting in 3,582, 2,062 and 6,016 test images respectively. We use ground truth object names and localise those in a given image. We report numbers on the PVOC 2007, COCO, and LVIS evaluation set. The mean Intersection over Union (IoU) is reported quantifying the overlap between the true and predicted bounding box. We report this metric for all bounding boxes and additionally for medium and large bounding box sizes only. A bounding box is considered large if it is over $96\times96$ pixels, and medium if between $32\times32$ and $96\times96$ pixels. We keep OpenFlamingo and BLIP-2 in its native form, which uses image resolutions of $224$, making it particularly difficult to localise small objects. For all experiments, we use the 3B parameter version with the instruction-tuned LLM of OpenFlamingo and the OPT 2.7B parameter version of BLIP-2. 

\paragraph{Implementation details.} The PIN module starts of from a 1D sinusoidal embedding~\cite{vaswani2017attention} with $64$ dimensions. From there a two-layer Multi-Layer-Perceptron is applied, each consisting of a fully connected (FC) layer, Layer Norm\cite{ba_layer_norm} and SwiGLU~\cite{swiglu}. Lastly, a final FC layer is added to match the target vision encoder embedding dimension of $1024$. The parameters of the PIN module are optimized with Adam~\cite{kingma2014adam} with a learning rate of $ 10^{-3}$. We train our PIN module on 2 $\times$ A6000 GPU for around two days. Overall, our PIN module consists of only around 1.2M parameters, \ie around 0.04\% of the VLM's size of 3B. 

\paragraph{Synthetic dataset details.} We follow X-Paste~\cite{xpaste} to create our synthetic dataset using Stable Diffusion~\cite{rombach_stable_diffusion} version 1 generating 60 samples for each category in LVIS~\cite{lvis} resulting in around 70k object images. We exclude all categories overlapping with COCO~\cite{coco} and PVOC~\cite{pvoc} for training. For the background, we use images from the BG20-k\cite{bg20k} dataset on which we paste the objects. Following X-Paste's filtering procedure, we exclude all classes with less than $\leq 20$ images remaining per class, as these classes might not be well-generated. For our composition function, we set the maximum allowed overlap to $o_{\text{max}}{=}0.5$, the number of images \( a_{\text{max}} {=} 3\), $r{=}r_{\text{orig}}$, \( s_{\text{min}} {=} [0.3, 0.2, 0.1] \) and \( s_{\text{max}} {=} [1.0, 1.0, 1.0] \), for up to three objects respectively.

\subsection{Quantitative Results}
\label{sub:Quantitative_Results}
\begin{figure*}[t!] \centering
    \includegraphics[width=\textwidth]{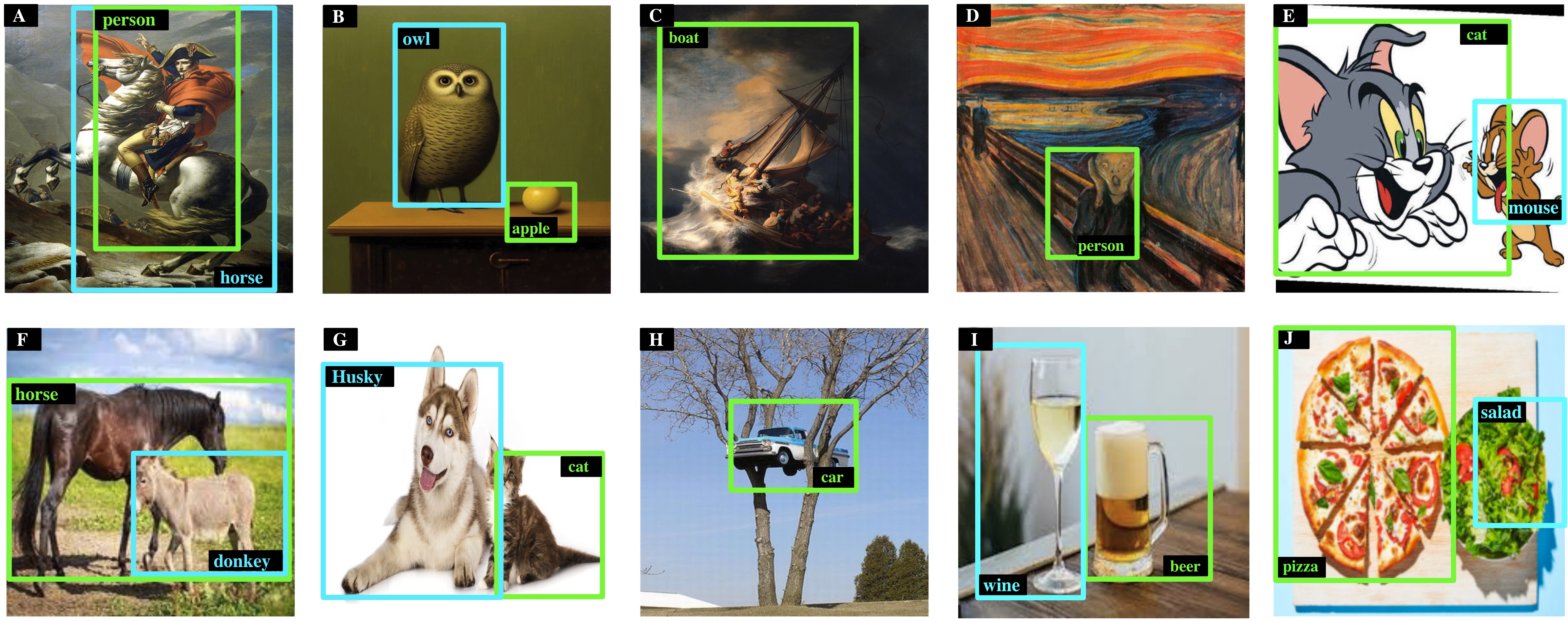}
    \caption{Localisation on a wide range of image types ranging from paintings, and comics to unique scenarios. Despite the varying image content, enhancing the OpenFlamingo caption-based VLM with our PIN shows strong localisation abilities.} \label{fig:ADE}
\vspace{-0.5cm}

\end{figure*}

\begin{figure*}[t!]
    \centering
    \includegraphics[width=\textwidth]{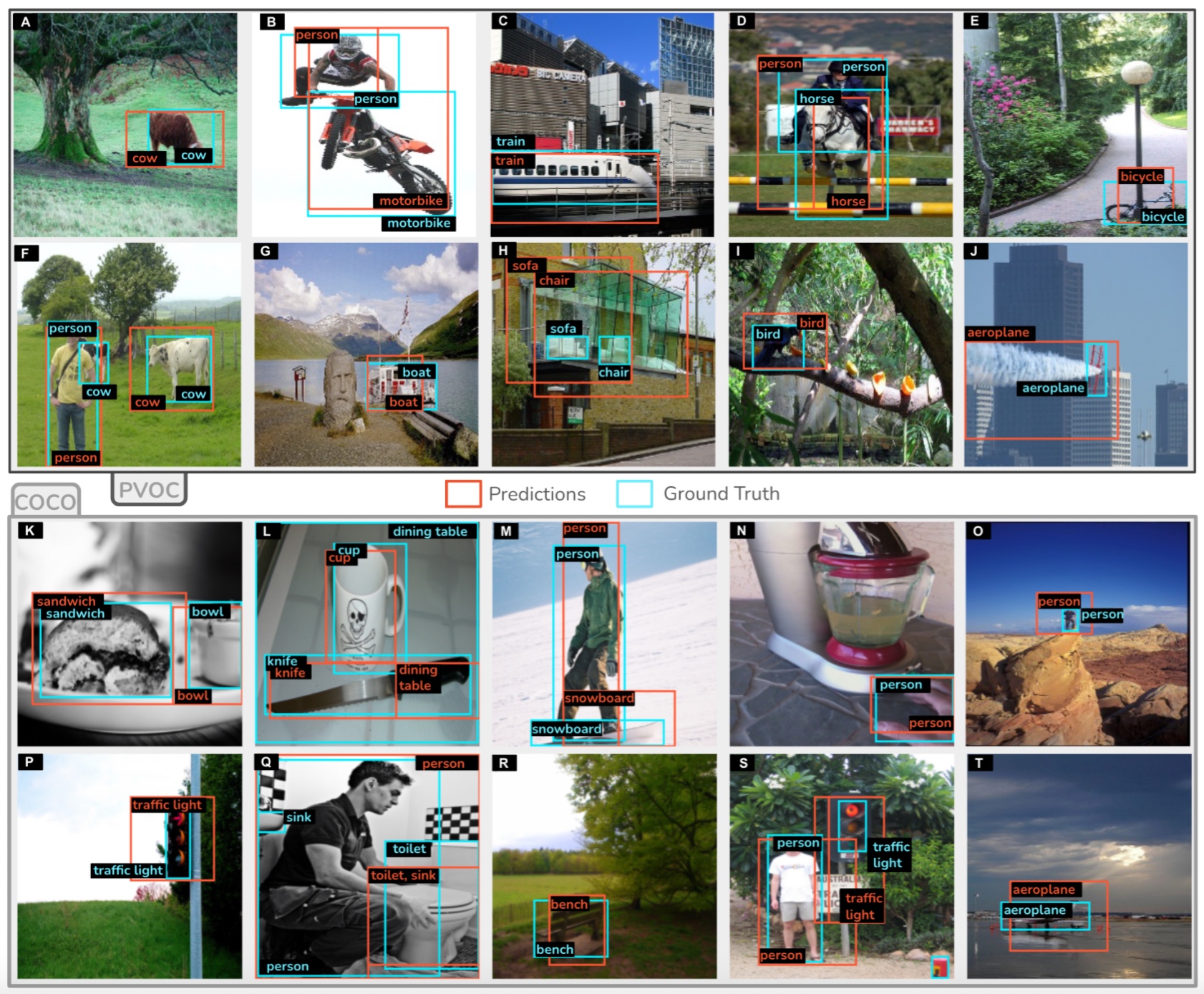}
    \caption{Object localisation results on PVOC~\cite{pvoc} and COCO~\cite{coco}. The PIN module unlocks spatial localisation in the caption-based OpenFlamingo~\cite{openflamingo_awadalla} VLM. 
    }
    \label{fig:coco_pvoc}
\vspace{-0.4cm}
\end{figure*}

\paragraph{Baselines.}
For comparison, we use OpenFlamingo's in-context learning version, configured with variable numbers of context images. To account for performance variation due to context image selection being sampled randomly, we execute each setup ten times and report the average and standard deviation. BLIP-2 is not able to do in-context learning due to the lack of interleaved image-text training data~\cite{blip2}. We select bounding boxes randomly from context images as a baseline to assess the in-context learning abilities. Additionally, we compare against other Parameter-Efficient Fine Tuning (PEFT) methods such as CoOp~\cite{zhou2022coop}, using the strongest version with adding 16 learnable tokens to the input to the LLM. In addition, we append 100 learnable tokens in the spirit of Visual Prompt Learning (VPT)~\cite{VPT_belongie} to either the vision encoder $\phi$ or the Fusion network $F$ (the same location where PIN is added). We also evaluate our method against finetuning the ViT vision encoder $\phi_V$ using LoRA~\cite{lora} with $\alpha$=16 and r=16.

\paragraph{Localisation on PVOC, COCO, and LVIS.}
From the results in Tab.~\ref{tab:method_comparison}, we first observe that our introduced PIN, when combined with OpenFlamingo, surpasses both the raw and the in-context learning versions of OpenFlamingo across all evaluated metrics, considerably. In particular, compared to the best OpenFlamingo in-context learning version, we improve in mIoU by a factor of $2\times$ on PVOC and a factor of $3\times$ on COCO. Notably, the PIN module achieves this without any exposure to COCO or PVOC classes during training, in contrast to the few-shot nature of in-context learning. The raw zero-shot OpenFlamingo variant fails to generate any meaningful bounding boxes, as visualized in Fig.~\ref{fig:study}. We observe that the random bounding box selector consistently performs better than the OpenFlamingo in-context learning version. This demonstrates that OpenFlamingo cannot leverage the positional information given by in-context bounding boxes to generate plausible bounding boxes for the query samples.  

Furthermore, we also compare the adapted VLM with PIN for OpenFlamingo against other Parameter-Efficient Fine-Tuning (PEFT) methods. First, we observe low performance of CoOp. This is primarily because of the lack of spatial positional information in CoOp's adaptation. OpenFlamingo employs a perceiver resampler as a fusion network, which removes most positional information during caption-based pretraining. Thus, the CoOp adaption struggles to solve the localisation task. In contrast, our PIN outperforms this baseline considerably, as it can add positional information directly to the vision embedding during adaption. We also compare against a different PEFT baseline which follows Visual Prompt Tuning (VPT)~\cite{VPT_belongie}, adding 100 learnable tokens to either the vision encoder $\phi_V$ or fusion network $F$. PIN outperforms the VPT baseline applied to the fusion network considerably and also the one applied to the vision encoder $\phi_V$, especially for medium-sized bounding boxes (IoU$_M$). These findings demonstrate that PIN better incorporates positional information into the pretrained VLM. In addition, we also show PIN's necessity by comparing it against finetuning the vision encoder $\phi_V$ with LoRA~\cite{lora}. PIN slightly outperforms the strong LoRA baseline while having 5$\times$ fewer parameters. We observe that the LoRA-adapted VLM can nearly perfectly solve our synthetic training examples, overfitting potentially to synthetic data artifacts. In contrast, PIN utilizes the strong concepts learned in the ViT without changing its weights, thus excluding the possibility of overfitting to synthetic data artifacts. We can also confirm the effectiveness of PIN on BLIP-2, outperforming again the other PEFT baselines. These findings demonstrate that PIN can effectively unlock localisation abilities in various VLMs beyond OpenFlamingo.
\input{tables/refcoco}

\begin{figure*}[t!]
    \centering
    \includegraphics[width=\textwidth]{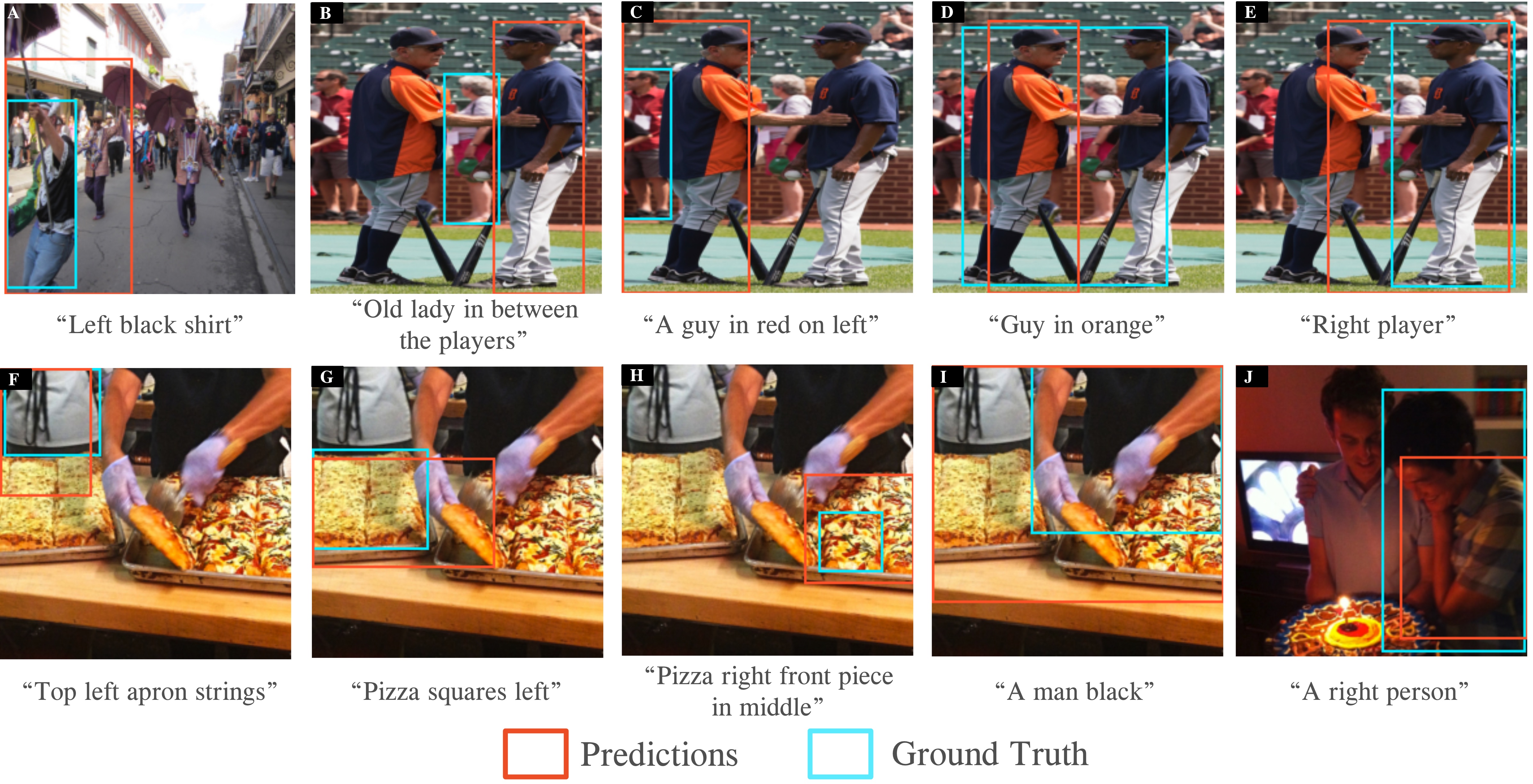}
    \caption{Zero-shot visual grounding results on RefCOCO~\cite{refcoco} of PIN with the OpenFlamingo~\cite{openflamingo_awadalla} VLM. The adapted VLM struggles with more complex scenarios(B and C), yet, it effectively handles simpler cases (F, G, H, J).}
    \label{fig:refcoco}
\end{figure*}

\paragraph{Grounding on RefCOCO.}
We also evaluate PIN on RefCOCO~\cite{refcoco} Test-A split in a zero-shot manner, paving a new way for reporting model performance \textit{without using any of its annotated training data}. To this end, we extend our synthetic dataset with positional expressions like `left apple', `monkey on the right' \etc. With this simplistic setup, we achieve $26.4$ P@$0.3$, indicating decent grounding abilities, compared to only $3.7$ for the in-context learning Flamingo baseline. Extending our synthetic data with referral expression improves results considerably, by a factor of nearly 2. In Fig.~\ref{fig:refcoco} we visualized our grounding predictions for RefCOCO. A limiting factor is the rather small 1B parameter LLM in OpenFlamingo, having trouble understanding more complex and longer referrals. 

\subsection{Qualitative Results}
\label{sub:Qualitative Results}

\paragraph{Localisation on diverse images.} We also explore the object localisation abilities of our adapted VLM on a wide range of images, encompassing various domains such as comics and paintings, as illustrated in Fig.~\ref{fig:ADE}. Notably, our method demonstrates robust performance in localising distinct characters and objects, even amidst significant domain variations. For instance, it successfully identifies the cat and mouse in a comic image (Fig.~\ref{fig:ADE}E) and accurately locates the person in a painting (Fig.~\ref{fig:ADE}D), as well as the owl and apple in another (Fig.~\ref{fig:ADE}B). Additionally, our VLM showcases its ability to differentiate between closely related objects. This is evident in its distinguishing between a donkey and a horse (Fig.~\ref{fig:ADE}F), as well as between a glass of wine and a glass of beer (Fig.~\ref{fig:ADE}I). These observations lead us to conclude that our adapted VLM not only excels in localising objects across varied image types but also retains the strong zero-shot capabilities typical of caption-based VLMs. 

\paragraph{Localisation on PVOC and COCO.} The adapted VLM accurately localises objects of different sizes, as demonstrated in Fig.~\ref{fig:coco_pvoc}. 
\textit{Variety in object sizes:} It identifies both large (person in Fig.~\ref{fig:coco_pvoc}Q) and small objects (bird in Fig.~\ref{fig:coco_pvoc}I; person in Fig.~\ref{fig:coco_pvoc}O). 
\textit{Variety in object locations:} We also find that the enhanced VLM localises objects at various locations in an image, \eg boxes near the bottom (Fig.~\ref{fig:coco_pvoc}C,E), top (Fig.~\ref{fig:coco_pvoc}B,M), left (Fig.~\ref{fig:coco_pvoc}F,R) and right (Fig.~\ref{fig:coco_pvoc}E,N).
\textit{Crowded and overlapping:} Additionally, our model effectively manages more complex situations such as more crowded scenes (train in Fig.~\ref{fig:coco_pvoc}C), partial occlusions (person riding a horse in Fig.~\ref{fig:coco_pvoc}D). 
\textit{Multi-object:} Our method is capable of localising multiple objects within a single image, demonstrating its ability to recognize more than just the most salient object. This can be seen \eg in Fig.~\ref{fig:coco_pvoc}Q for the person and toilet and in Fig.~\ref{fig:coco_pvoc}B for the person and the motorbike.
Yet, the adapted model struggles with more confusing scenes yielding more loose bounding box predictions like the trail of the aeroplane in Fig.~\ref{fig:coco_pvoc}J. Similarly, for small bounding boxes, our approach cannot locate objects very precisely, \eg the sofa and chair in Fig.~\ref{fig:coco_pvoc}H or sink in Fig.~\ref{fig:coco_pvoc}Q. Overall, we conclude that the model can extend its zero-shot abilities to the object localisation task. In Fig.~\ref{fig:blip}, we visualize results with the BLIP-2 VLM.

\subsection{Ablations}

\input{tables/higher_res}

\paragraph{Generalization of synthetic data.}
In Tab.~\ref{tab:higher_res} (\textit{Generalization}), we delve deeper into the choice of training data on the zero-shot abilities of our PIN module. For that, we compare training PIN on either the COCO datasets or using the synthetic data in which all COCO and PVOC categories are excluded. As expected, we observe better performance for the PIN trained on COCO and evaluated on COCO. However, we observe equivalent performance when analyzing their generalization abilities to PVOC. From that, we conclude that synthetic data serves as a viable solution to adapt pretrained VLMs for object localisation while preserving their generalization capabilities.

\begin{figure*}[t] \centering
    \includegraphics[width=\textwidth]{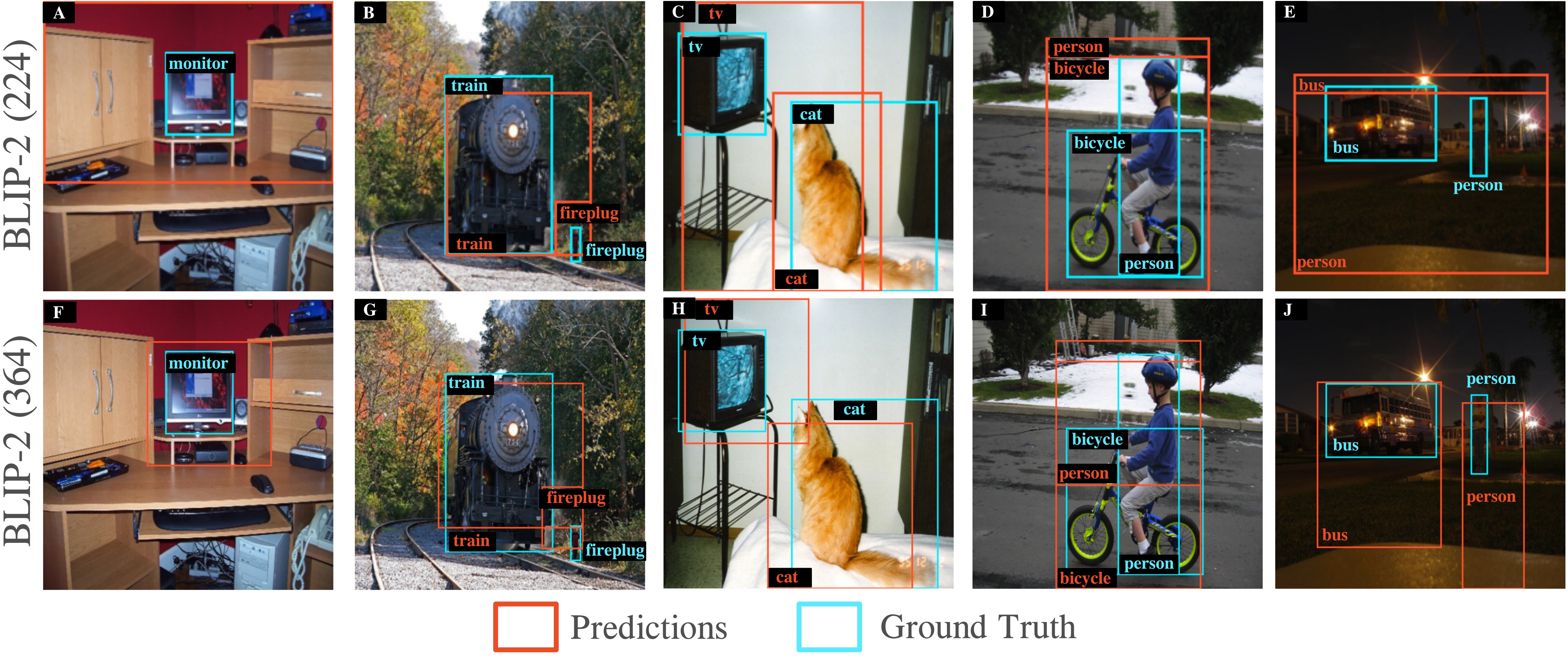}
    \caption{Object localisation results with BLIP-2~\cite{blip2} on 224$\times$224, BLIP-2 (224), image resolution and 364$\times$364, BLIP-2 (364), on PVOC~\cite{pvoc}. The PIN trained with the higher image resolution BLIP-2 version is able to predict more accurate bounding boxes.} 
    \label{fig:blip}
\vspace{-0.2cm}
\end{figure*}

\paragraph{Higher image resolution.}
In Tab.~\ref{tab:higher_res} (\textit{Higher Resolution}), we analyze the impact of using higher image resolutions on the performance of PIN. All OpenFlamingo models are pretrained on a resolution of $224\times 224$. To circumvent that, we extrapolate the frozen positional embeddings of the ViT, allowing our PIN to be trained at a resolution of $448\times 448$. As expected, this leads to an improvement across all IoU metrics, particularly for medium-sized bounding boxes (mIoU$_M$). We scaled the size of the bounding box for medium M, and large L according to the increase in scale of the image resolution. Most VLMs of BLIP-2 are trained on an image resolution of $224\times 224$, yet, caption finetuned VLMs are available on $364 \times 364$ image resolution. We visually compare the difference in Fig.~\ref{fig:blip} and observe tighter bounding boxes with the higher resolution VLM. Training PIN on a higher BLIP-2 resolution results in similar IoU improvements as for OpenFlamingo.


\paragraph{Impact of PIN on VLM's general abilities.} We analyze the impact of applying PIN on the general abilities of the VLM using the VQAv2~\cite{VQA2} dataset. The base performance of OpenFlamingo is $44.1\%$ when inserting PIN, the performance reduces to \(34.3\%\), yet it does not compromise the VLM. Moreover, we compare this to the VLM adapted with the finetuned vision encoder. We observe a bigger reduction in performance with \(33.4\%\). In addition, our PIN can be easily deactivated, thereby retaining the general VLM abilities, a flexibility not possible when finetuning the ViT.


\paragraph{Amount of objects to paste.} Lastly, we evaluate the maximum number of objects, denoted as \( a_{\text{max}} \), that are pasted onto the background for each image. Separate models are trained for 1, 2, 3, 4, and 5 allowed objects per image. The results are shown in Tab.~\ref{tab:ablation1} and the mIoU on the COCO dataset is reported for a maximum of 3, 4, and 5 objects per image. We observe a decrease in performance when too few objects are pasted during training (mIoU$_{\leq 3}$ of 0.24 vs. 0.35) as the VLM only focuses on the most salient object. Alternatively, pasting too many objects also decreases performance, especially for mIoU$_{\leq 5}$. With \( a_{\text{max}} {=} 3 \) we strike a good balance between these two extremes, yielding the best accuracies across all mIoU values.

\input{tables/ablation2}





%% file: tables/refcoco.tex
\begin{table}[t]
\centering
\begin{tabular}{lc}
\toprule
OpenFlamingo\cite{openflamingo_awadalla} & P@0.3 \\
\midrule
+ Raw & 0 \\
+ In-context learning & 3.7  \\
+ PIN w/o positional referral& 14.1  \\
+ PIN w/ positional referral & 26.4  \\
\bottomrule
\end{tabular}

\caption{Evaluation on RefCOCO \cite{refcoco} Test-A. PIN shows decent grounding abilities without using any annotated training data, outperforming the in-context learning Flamingo baseline. Extending our synthetic dataset with positional referrals improves performance considerably.}
\label{tab:performance_comparison}
\vspace{-0.8cm}
\end{table}

%% file: tables/higher_res.tex
\begin{table}[t]
\centering
\resizebox{\columnwidth}{!}{%
\begin{tabular}{ll ccc | ccc }
\toprule
&\multirow{2}{*}{Method} & \multicolumn{3}{c}{$\text{PVOC}_{\leq 3 \text{ Objects}}$ } & \multicolumn{3}{c}{$\text{COCO}_{\leq 3 \text{ Objects}}$ } \\
&& mIoU & $\text{mIoU}_M$ & $\text{mIoU}_L$ & mIoU & $\text{mIoU}_M$ &  $\text{mIoU}_L$ \\
\midrule
& \textit{Generalization} \\
\parbox[t]{0mm}{\multirow{2}{*}{\rotatebox[origin=t]{90}{OpenFlamingo}}}& \hspace{2mm} \includegraphics[width=0.9em]{figures/pin_emoji.png } PIN (COCO) &\B 0.45 &\B 0.27 &\B 0.63 & \B 0.39 &\B 0.31 & \B 0.62 \\
& \hspace{2mm} \includegraphics[width=0.9em]{figures/pin_emoji.png } PIN (Synth.) &\B 0.45 &\B 0.27 & 0.62 & 0.35 &0.26 & 0.59 \\
\cmidrule{2-8}
& \textit{Higher Resolution} \\
& \hspace{2mm} \includegraphics[width=0.9em]{figures/pin_emoji.png } PIN (224) & 0.45 &0.27 & 0.62 & 0.35 & 0.26 & \B 0.59 \\
& \hspace{2mm} \includegraphics[width=0.9em]{figures/pin_emoji.png } PIN (448) & \B 0.47 & \B 0.30 & \B 0.65 & \B 0.37 &\B 0.29 &\B 0.59 \\
\midrule

\parbox[t]{0mm}{\multirow{2}{*}{\rotatebox[origin=t]{90}{BLIP2}}}& \hspace{2mm} 
\includegraphics[width=0.9em]{figures/pin_emoji.png } PIN (224) & 0.44  & 0.24 & 0.63 & 0.34& 0.22 & 0.60 \\
& \hspace{2mm} \includegraphics[width=0.9em]{figures/pin_emoji.png } PIN (364) & \B 0.47 & \B 0.27 &\B 0.66 & \B0.37 &\B 0.26 &\B 0.62 \\
\bottomrule

\end{tabular}
}
\caption{Ablating the image resolution and the choice of synthetic training data for PIN.}
\label{tab:higher_res}
\vspace{-0.4cm}
\end{table}

%% file: tables/ablation2.tex
\begin{table}[t]
\centering
\resizebox{0.42\textwidth}{!}{%
\begin{tabular}{ cccc }
\toprule
\# pasted objects & mIoU$_{\leq 3}$ & mIoU$_{\leq 4}$ & mIoU$_{\leq 5}$\\
\midrule
$\leq$ 2 & 0.24 & 0.21 & 0.19 \\
$\leq$ 3 &  \B 0.35 &  \B 0.31 &  \B 0.29 \\
$\leq$ 4 & 0.35 & 0.30 & 0.28 \\
$\leq$ 5 & 0.34 & 0.30 & 0.27 \\

\bottomrule
\end{tabular}
}
\caption{Ablation on the number of objects being pasted during training on our synthetic data evaluated on COCO. Pasting with 1-3 objects works best across all mIoU scores.}
\label{tab:ablation1}
\vspace{-0.5cm}
\end{table}

%% file: 6_conclusions.tex
\section{Conclusion}%
\label{sec:Conclusion}
In this work, we introduced PIN, a lightweight module that enables object localisation capabilities in a frozen VLM. We first showed the limited object localisation abilities of caption-based VLMs. Subsequently, we verified that these capabilities were enabled with our PIN module on OpenFlamingo and BLIP-2. Our zero-shot results across PVOC and COCO, various image types, and objects demonstrate that the strong performance of caption-based VLMs can be transferred to localisation. 

\noindent \textbf{Limitations.} Owing to our simplistic training procedure and the caption-based pretraining focusing on big objects in relatively low-resolution images, our model struggles with generating tight bounding boxes, especially around smaller objects. As a \textit{no-bells-and-whistles} paper, we leave these challenges to future work.

\subsection*{Acknowledgement}
This work is financially supported by Qualcomm Technologies Inc., the University of Amsterdam, and the allowance Top consortia for Knowledge and Innovation
(TKIs) from the Netherlands Ministry of Economic Affairs and Climate Policy. 

%% file: 7_supp.tex



\section{Extended analysis of caption-based VLMs}
\label{sec:extended_study}
\begin{figure*}[!h]
    \centering
    \begin{subfigure}[b]{1\textwidth}
        \includegraphics[width=\textwidth]{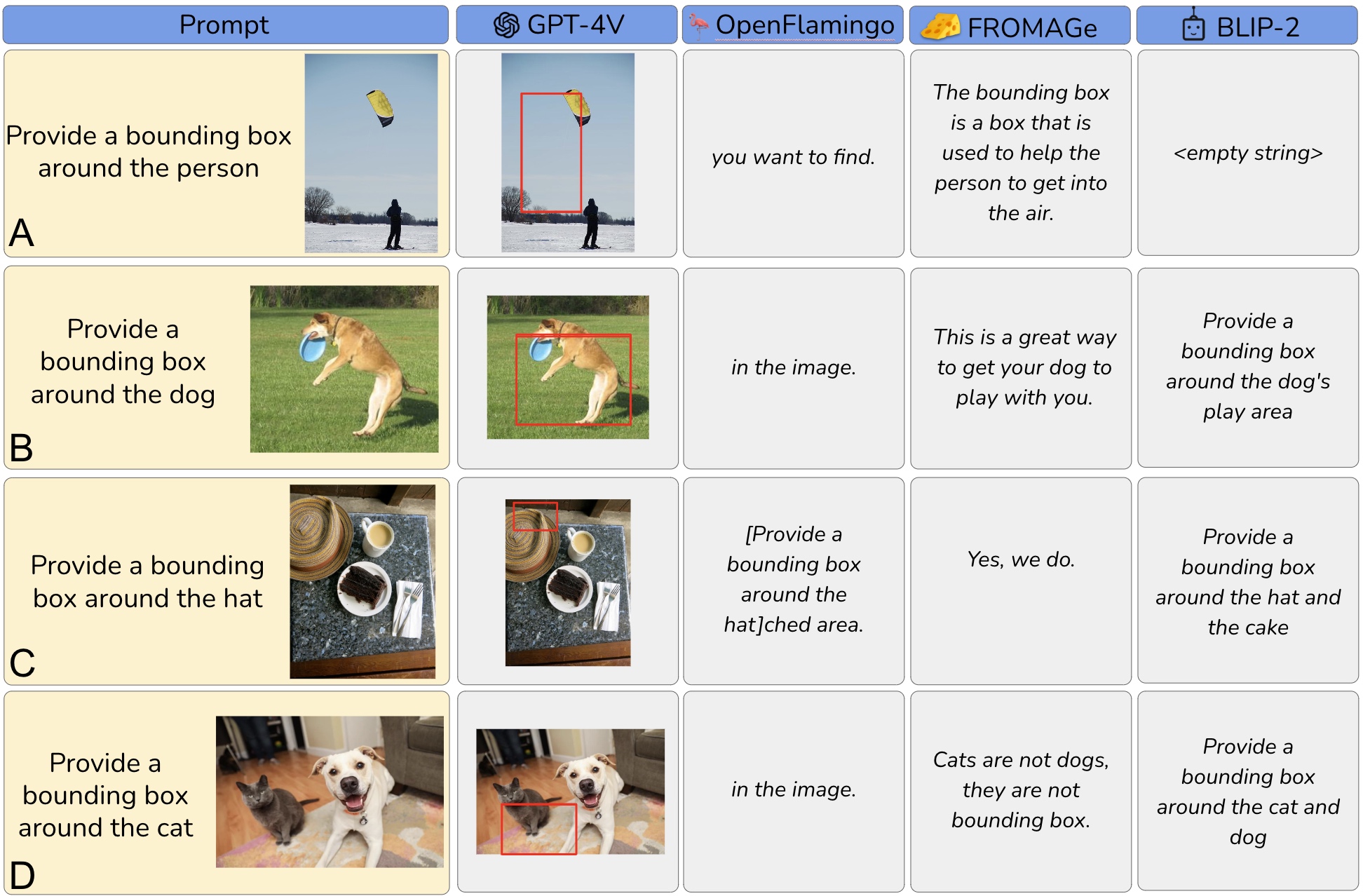}
    \end{subfigure}
    \hfill 
    \begin{subfigure}[b]{1\textwidth}
        \includegraphics[width=\textwidth]{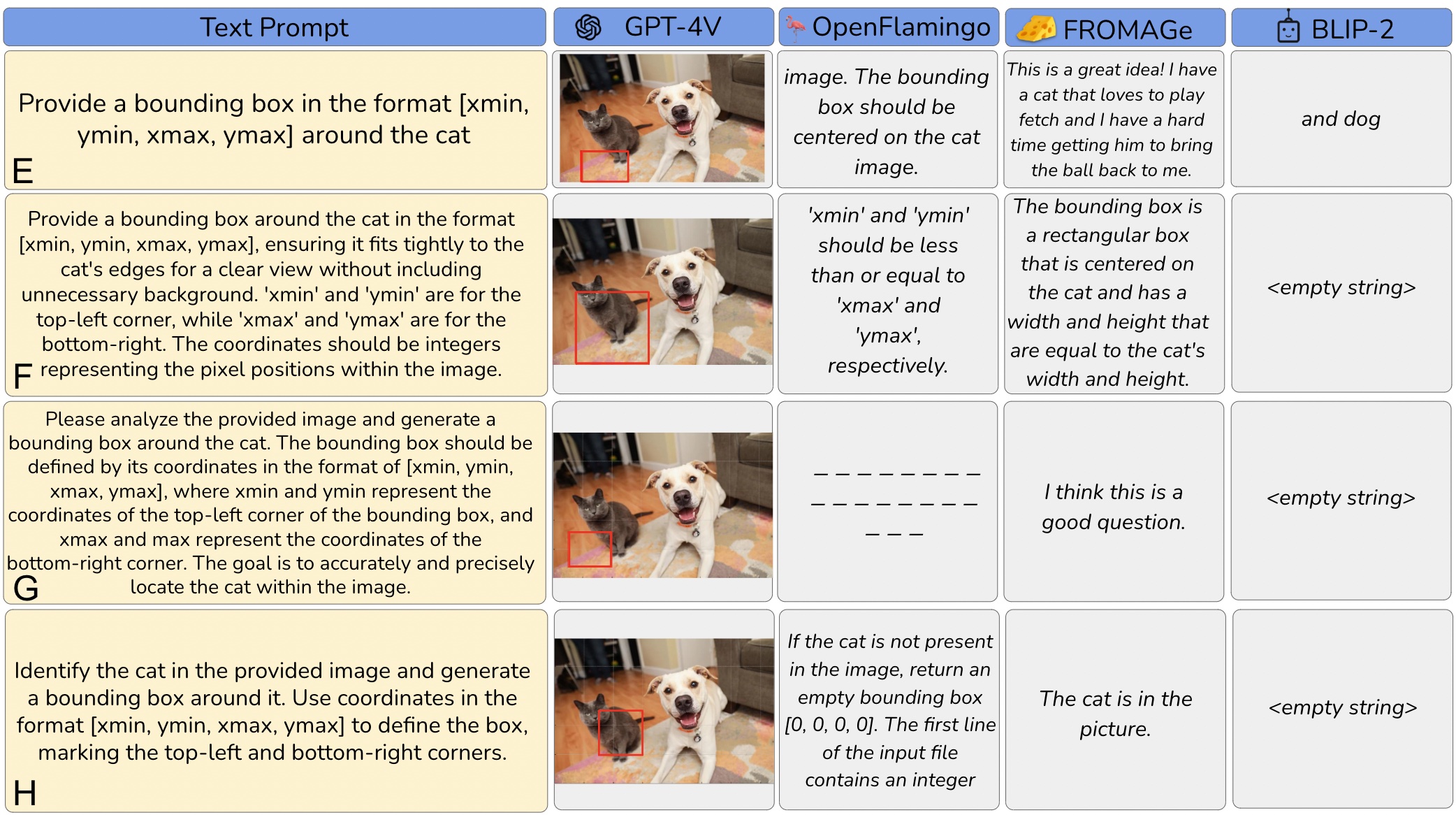}
    \end{subfigure}
    \caption{Analysis of localisation abilities of caption-based VLMs to provide a bounding box. A-D shows results with the same prompt on different sample images and E-H illustrates prompts with more instruction information on the same cat and dog image (D).} 
    \label{fig:bbox_study}
\end{figure*}

\begin{figure*}[!h]
    \centering
    \begin{subfigure}[b]{0.94\textwidth}
        \includegraphics[width=\textwidth]{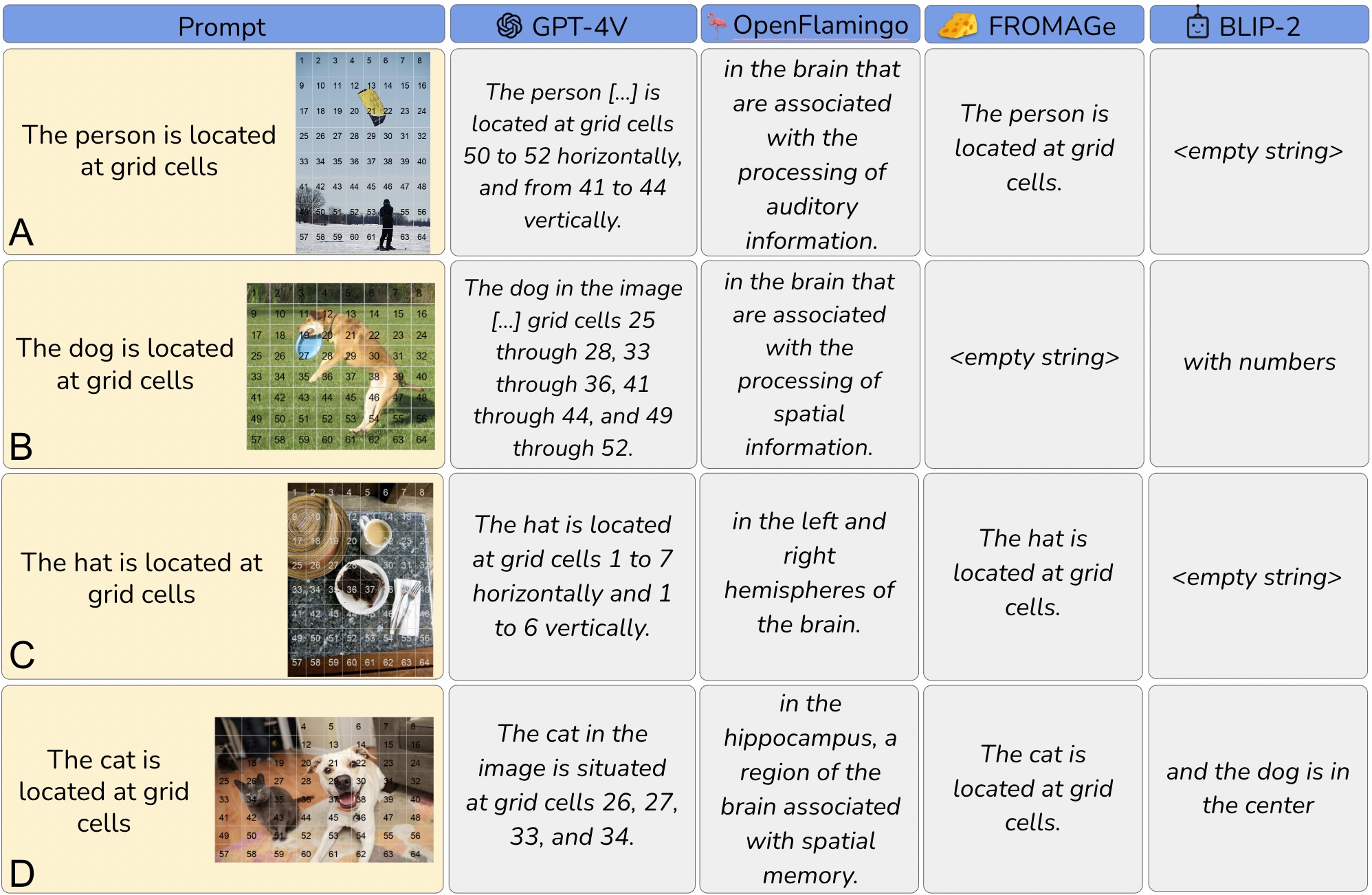}
    \end{subfigure}
    \hfill 
    \begin{subfigure}[b]{0.94\textwidth}
        \includegraphics[width=\textwidth]{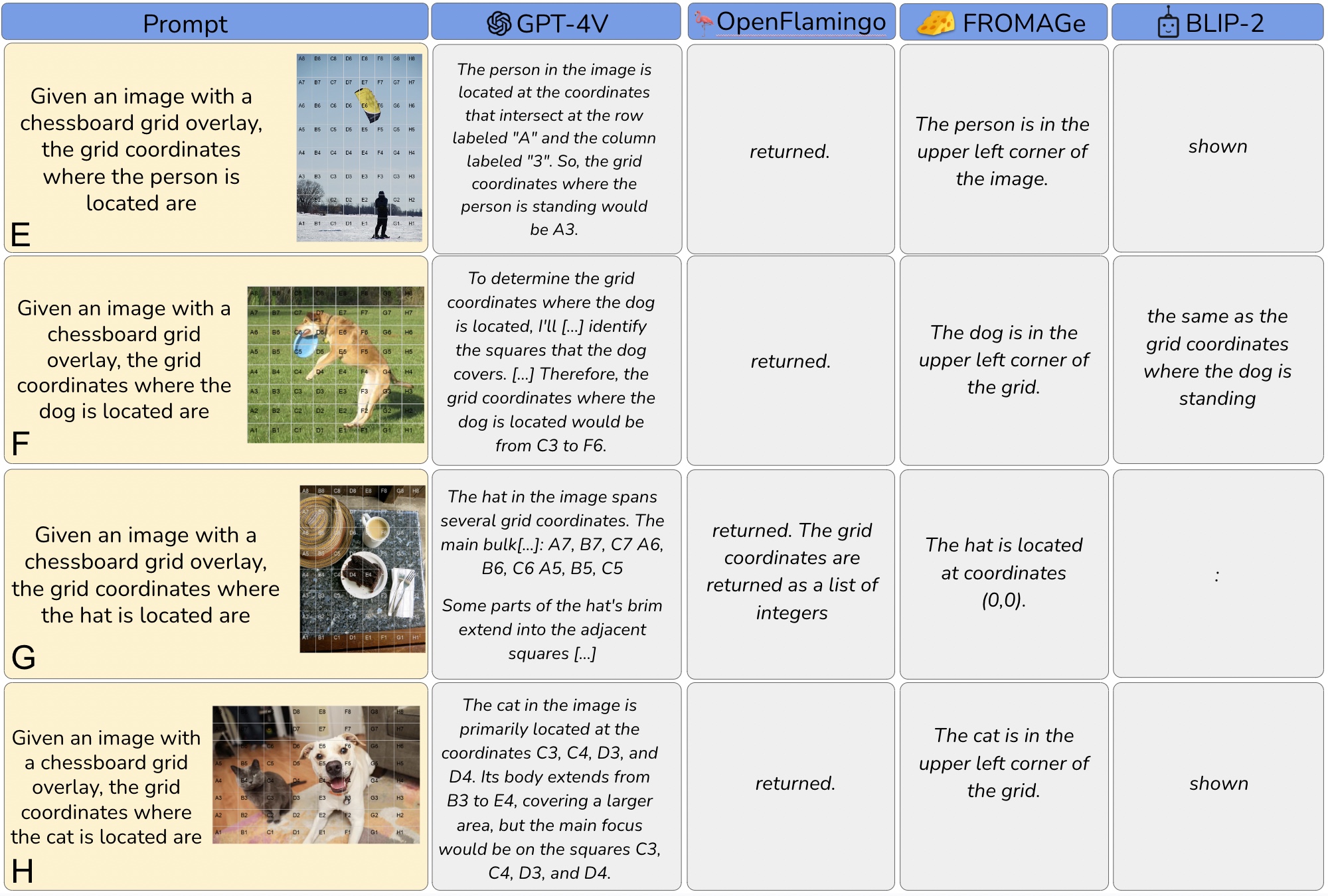}
    \end{subfigure}
    \caption{Analysis of grid-based localisation of caption-based VLMs. A-D shows results with a numbered grid overlaid on the image. while E-H shows findings with a checkerboard-style grid.} 
    \label{fig:grid_study} 
\end{figure*}

\begin{figure*}[!h]
    \centering
    \begin{subfigure}[b]{0.95\textwidth}
        \includegraphics[width=\textwidth]{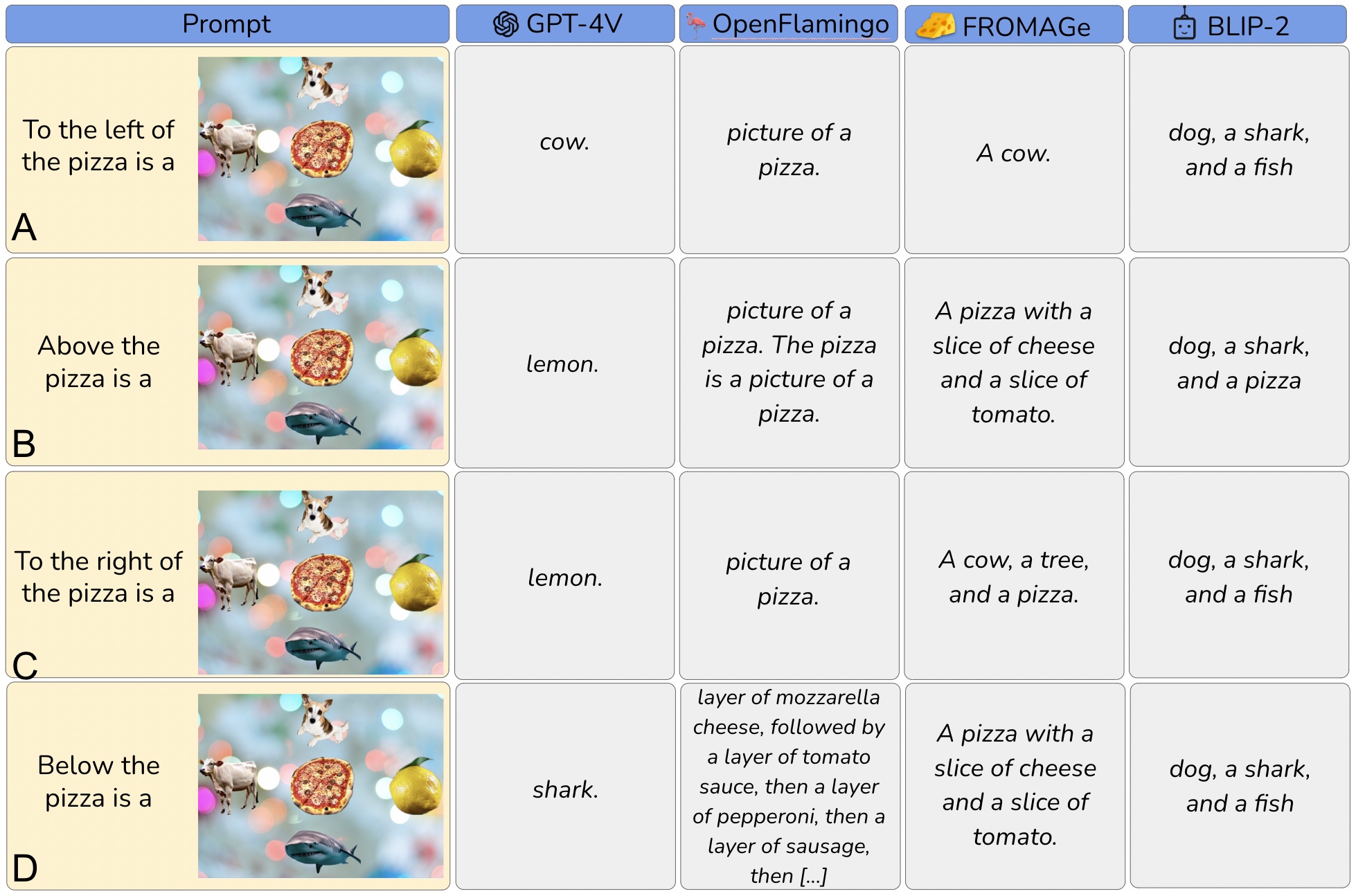}
    \end{subfigure}
    \hfill 
    \begin{subfigure}[b]{0.95\textwidth}
        \includegraphics[width=\textwidth]{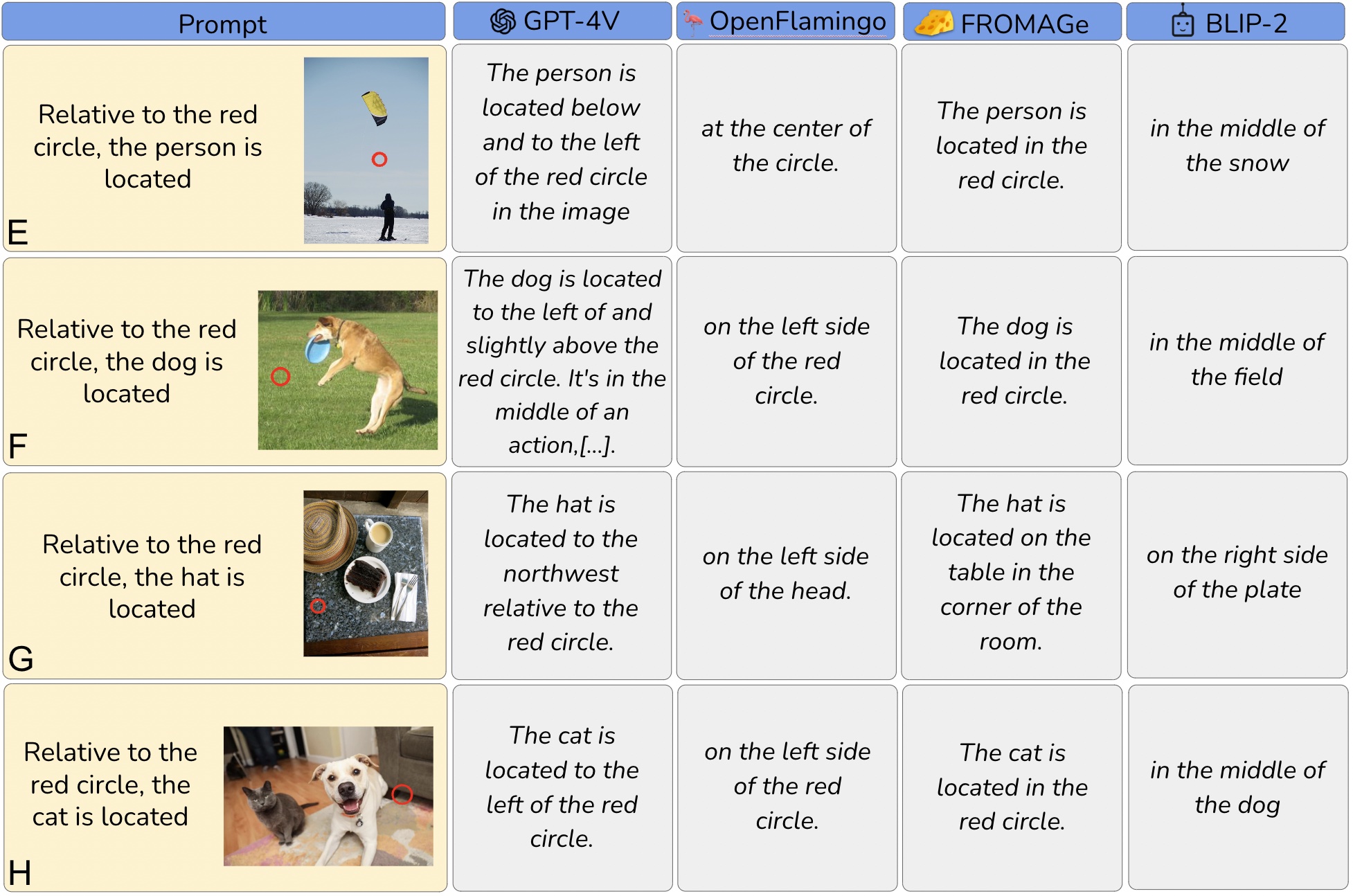}
    \end{subfigure}
    \caption{Analysis of relative position abilities of caption-based VLMs. In A-D, VLMs have to identify the object relative to the center one. In E-H, VLMs are tasked to provide the location relative to a red circle.
    }
    \label{fig:rel_position_study} 
\end{figure*}

This section broadens the scope of our analysis of the localisation abilities of caption-based Vision-Language Models (VLMs) from the main paper. Our goal is to assess a wider range of prompts on more sample images. The study employs the same collection of VLMs as before, namely:
\begin{itemize}
    \item GPT-4V~\cite{open_ai-gpt4}
    \item 7B version of BLIP-2~\cite{blip2}
    \item 9B version of Flamingo~\cite{openflamingo_awadalla, alayrac2022flamingo}
    \item Fromage~\cite{koh2023grounding}
\end{itemize}
Note that due to the undisclosed training data for \mbox{GPT-4V}~\cite{open_ai-gpt4}, we cannot rule out its exposure to supervised object localisation training. Our expanded analysis includes three prompt types, designed to test the VLMs' abilities in various aspects of spatial understanding and object localisation. The prompts cover a spectrum of challenges, from generating bounding boxes around a specified object (shown in Fig.~\ref{fig:bbox_study}) to performing grid-based localisation (illustrated in Fig.~\ref{fig:grid_study}) and determining relative positions (depicted in Fig.~\ref{fig:rel_position_study}).

\paragraph{Generate bounding box} 
Similar to the study in the main paper, we evaluate caption-based VLMs in their ability to generate a bounding box for the specified object. For this purpose, we applied the prompt from the main paper to more sample images, which are depicted at the top of Figure~\ref{fig:bbox_study}. Our observations indicate that only GPT-4V is capable of generating a bounding box that is approximately located near the object of interest, yet not with high precision; for example, the cat in Figure~\ref{fig:bbox_study}D. In contrast, all other VLMs, such as OpenFlamingo as shown in Figure~\ref{fig:bbox_study}B, complete the sentence with 'in the image,' without providing any bounding box informatiom. To further evaluate these VLMs, we added more detailed instructions to the prompt, such as `in the format of \([x_{\text{min}}, y_{\text{min}}, x_{\text{max}}, y_{\text{max}}]\)', which can be found in Figure~\ref{fig:bbox_study}E-H. We observe that even with more instructions, these VLMs are not able to provide any bounding box or positional information about the inquired object. 

\paragraph{Grid-based localisation} 
In this part, we evaluate the VLMs with a grid-based localisation task using two different grid styles. The first style uses a standard numbered grid (Fig.~\ref{fig:grid_study}A-D), while the second uses a chessboard-style grid (Fig.~\ref{fig:grid_study}E-H). In both cases, an 8x8 grid is overlaid on the images. The size of each grid cell varies to match the aspect ratio of the image. The goal is to evaluate the VLMs' ability to pinpoint the object location using the designated grid. We observe that only GPT4-V is able to list grid cells in its response, yet, for the numbered grid its response does not match the objects (e.g. the dog in Fig.~\ref{fig:grid_study}B), and for it only roughly matches the object (e.g. the cat in Fig.~\ref{fig:grid_study}H).
The other models generally fail to provide accurate or relevant coordinates in response to the grid-based prompts. Their responses are often off-task, with Flamingo providing unrelated continuations (such as 'cells […] of the brain'), Fromage repeating the prompt, and BLIP-2 sometimes not responding at all. This indicates a gap in these models' ability to understand and execute spatial tasks.

\paragraph{Relative position} 
Here, we evaluate the VLMs' relative position abilities. For that, we task the models to identify an object relative to a center object (Fig.~\ref{fig:rel_position_study}A-D). Therefore, we designed an artificial image with a pizza at the center, surrounded by a lemon to the left, a shark to the bottom, a cow to the right, and a dog above. We observe that BLIP-2 listed three random objects, regardless of the prompt. Fromage detects the objects to the left correctly Fig.~\ref{fig:rel_position_study}A, yet, all other directions are wrong. OpenFlamingo responses are only about the pizza ignoring the surrounding objects. GPT-4V does answers correctly for all directions except for the one above the pizza Fig.~\ref{fig:rel_position_study}B. 
We extend our study to ask VLMs how a specific object is placed relative to a red circle that is overlaid on the image (Fig.~\ref{fig:rel_position_study}E-H). This is inspired by \cite{red_circle_vedaldi} which showed that red circles can be used for VLMs to direct their attention to a specific region. We observe that Fromage and BLIP-2 are not able to provide any meaningful responses. Instead, often these VLMs try to describe the absolute position of the object e.g.for Fromage Fig.~\ref{fig:rel_position_study}C and BLIP-2 Fig.~\ref{fig:rel_position_study}B. OpenFlamingo answers give indeed relative positional information, yet, most often wrong and in 3 of 4 cases `on the left side'. Again, only GPT4-V is able to give roughly correct responses e.g. Fig.~\ref{fig:rel_position_study}D, yet, Fig.~\ref{fig:rel_position_study}A and C are partially and Fig.~\ref{fig:rel_position_study}B is completely wrong. From that, we conclude that caption-based VLMs struggle with solving relative positional tasks indicating a lack of spatial understanding on the relative placement of objects.

\paragraph{Summary}
The extended analysis of caption-based VLMs reveals limitations in their spatial understanding and object localisation abilities. Among all evaluated models, only GPT-4V managed to generate responses that partially met the task criteria.
Yet, due to the undisclosed training data for GPT-4V~\cite{open_ai-gpt4}, we cannot rule out its exposure to supervised object localisation training. 
Despite varying prompt complexities and image scenarios, all other VLMs consistently underperform in tasks requiring precise localisation and relative positioning. The study's findings underscore a gap in the current capabilities of caption-based VLMs, highlighting their struggles with accurately interpreting and responding to spatially-oriented tasks. This motivated us to design the PIN module to unlock localisation abilities in the caption-based VLM Flamingo.

\begin{figure*}[t] 
    \centering
    \includegraphics[width=\textwidth]{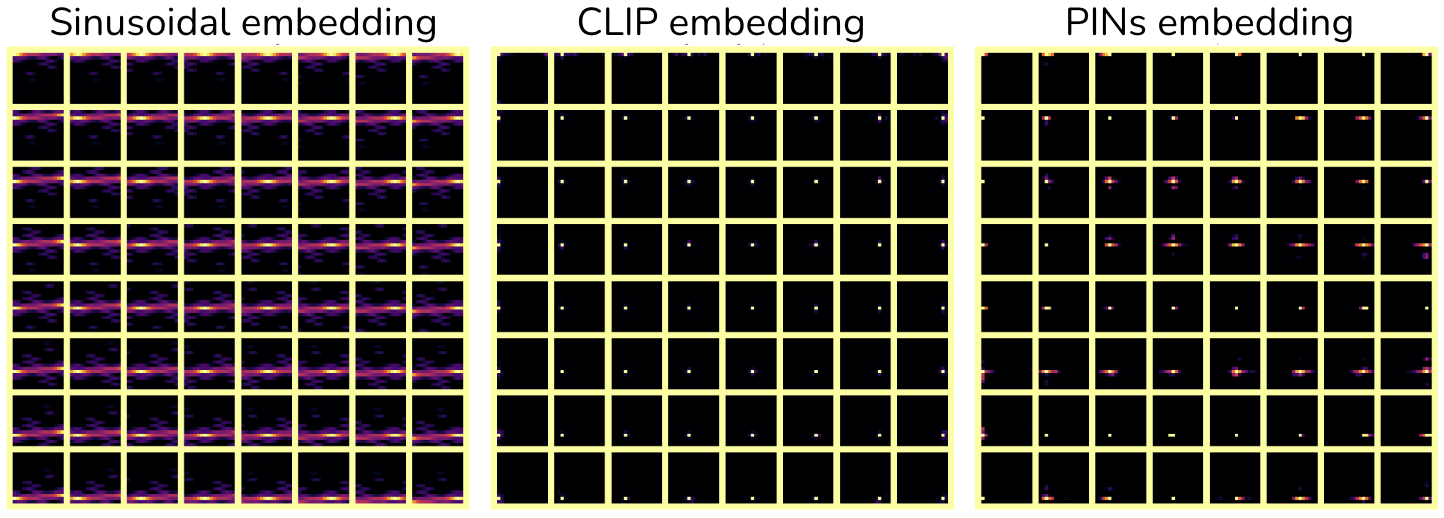} 
    \caption{Visualization of pair-wise similarities of the raw sinusoidal embedding,  the CLIP encoder's spatial embeddings and our learned PIN. Our embedding captures local positional information, making it effective for localisation.}
    \label{fig:vis_PIN}
\end{figure*}
\input{tables/ablation3}

\section{Additional ablations}

\paragraph{Visualizing $\pi$} In Fig.~\ref{fig:vis_PIN}, we present a visualization of our learned input-independent feature vector $\pi$ from the PIN module. Following ViT~\cite{vit}, we compute the cosine similarity for all pairings of the $16{\times}16$ patches. This results in a $16{\times}16$ grid visualization, where each cell shows the similarity between a specific patch with all other patches. For readability, we omitted every second patch, thus being a $8{\times}8$ plot. We also visualize the 1D sinusoidal embedding as it is the starting point for our PIN module. From Fig.~\ref{fig:vis_PIN}, we find that this embedding only obtains the highest similarities with itself and among patches in the same row, a characteristic feature of the sinusoidal embeddings. Conversely, our learned embedding $\pi$ demonstrates high similarity primarily within itself \textit{and} its immediate neighboring patches, an attribute advantageous for localisation tasks, highlighting the similarity among the spatial locations. We also visualize the similarity for the raw CLIP vision encoder embeddings by averaging the similarities over $50$ images. 
We observe that the embedding of the vision encoder does not contain any positional information as only one bright spot, the similarity with itself, can be found in each cell. In summary, our visualizations show that our learned embedding $\pi$ successfully captures local positional information, making it particularly effective for tasks like localisation.

\subsection{Depth of $\psi$}
In this ablation, we analyse the impact of varying the number of layers in the feed-forward neural network $\psi$ inside the PIN module. Table ~\ref{tab:ablation3} (a) - (c) displays the results. Increasing the layer quantity results in a rise in parameters, advancing from 0.6M for one layer to 1.2M for two layers, and reaching 2.3M for three layers.
We find that the optimal number of layers in $\psi$ is 2, as evidenced by the highest mIoU scores across all categories. The findings indicate that a few learnable parameters are sufficient, aligning with the input-agnostic characteristics of the PIN module.

\subsection{Sinusoidal vs learned}
We investigate the effectiveness of the sinusoidal embedding~\cite{bert} and compare it against a learned variant. As shown in Table ~\ref{tab:ablation3} (d) - (e), both types of embeddings yield similar performance, with no significant difference in mIoU scores. 
Our goal is to incorporate spatial information into the VLM, for which the sinusoidal embedding is ideally suited. Its performance matches that of the learned version, which in theory provides greater adaptability and capacity for the model. Thus, the sinusoidal embedding with no learnable parameters is the optimal choice for our PIN module due to its efficiency and effectiveness in this context.

\subsection{Choice of background} 
We ablate the choice of background images for our synthetic data generation. To this end, we compare the BG-20k~\cite{bg20k} by using plain white background images on COCO in Tab.~\ref{tab:ablation2} rows (a-b). We observe a strong performance decrease in terms of IoU with white backgrounds, especially for medium-sized bounding boxes. 
We conjecture that the more realistic images in BG-20k contribute to a more robust spatial embedding $\pi$, enhancing localisation performance.

\input{tables/ablations1}
\subsection{Overlap between objects}
Lastly, we evaluate the effect of allowing for overlap $o_{\text{max}}$ between pasted objects during training on our synthetic generated data on COCO. We compare two settings of no-overlap $o_{\text{max}}{=}0.5$ in Tab.~\ref{tab:ablation2}, rows (c-d). We find that by creating more realistic generations by allowing for overlapping pasted objects, we obtain slightly better localisation performance, indicating a better learned PIN module.

\section{Additional qualitative results}
\subsection{Visualizations on LVIS}
\begin{figure*}[t] \centering
    \includegraphics[width=\textwidth]{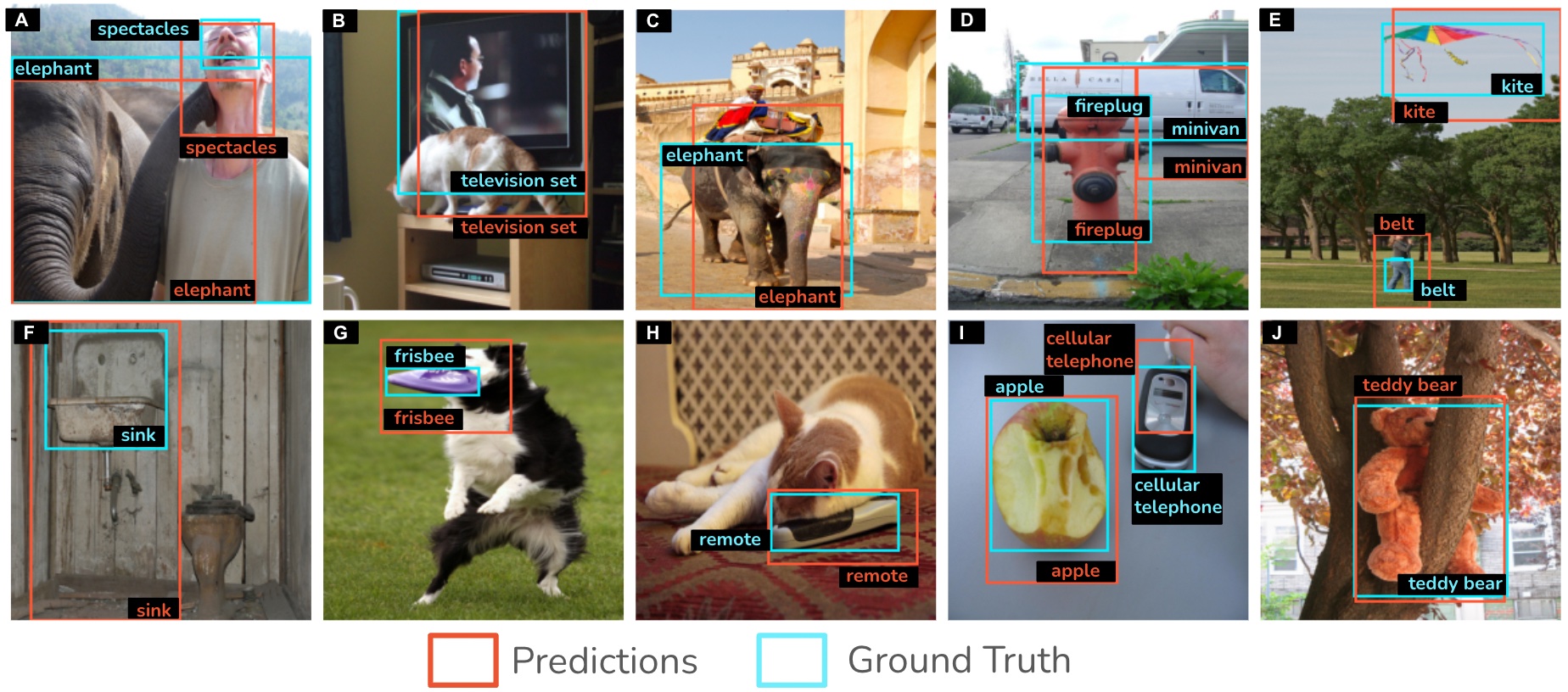}
    \caption{Object localisation results on LVIS~\cite{lvis} with the OpenFlamingo VLM.} \label{fig:lvis}
\end{figure*}

Our adapted VLM demonstrates effective object localisation also on LVIS~\cite{lvis} as demonstrated in Fig.~\ref{fig:lvis}. Our model can localise multiple objects within a single image, as illustrated in Fig.~\ref{fig:lvis}A, D, E, and I. It also effectively identifies objects in unusual settings, such as a teddy bear in a tree (Fig.~\ref{fig:lvis}J) and a remote under a cat (Fig.~\ref{fig:lvis}H). These examples support the conclusion that our model extends its zero-shot capabilities to the task of object localisation.

\subsection{Zero-shot visualizations on synthetic data} 
\begin{figure*}[t] \centering
    \includegraphics[width=\textwidth]{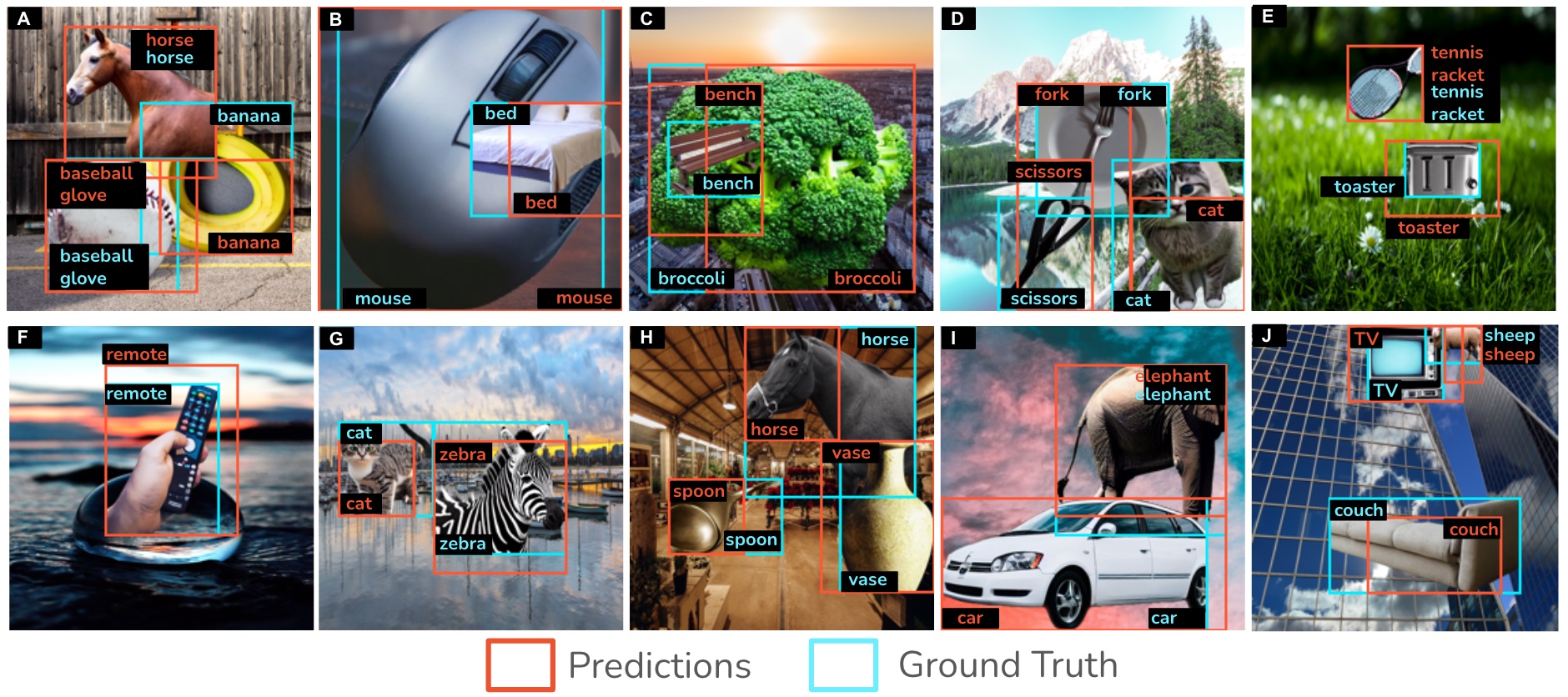}
    \caption{Zero-shot object localisation results on our synthetic data with the OpenFlamingo VLM.} \label{fig:synthetic_zero_shot}
\end{figure*}

In Fig.~\ref{fig:synthetic_zero_shot}, we demonstrate the zero-shot localisation capabilities of our VLM on our synthetic generated data. This visualization showcases the model's ability to accurately identify and localise multiple objects within an image, even in scenarios where pixel boundaries are not distinctly defined. 

\subsection{Visualizations of failure cases}
\begin{figure*}[t] \centering
    \includegraphics[width=\textwidth]{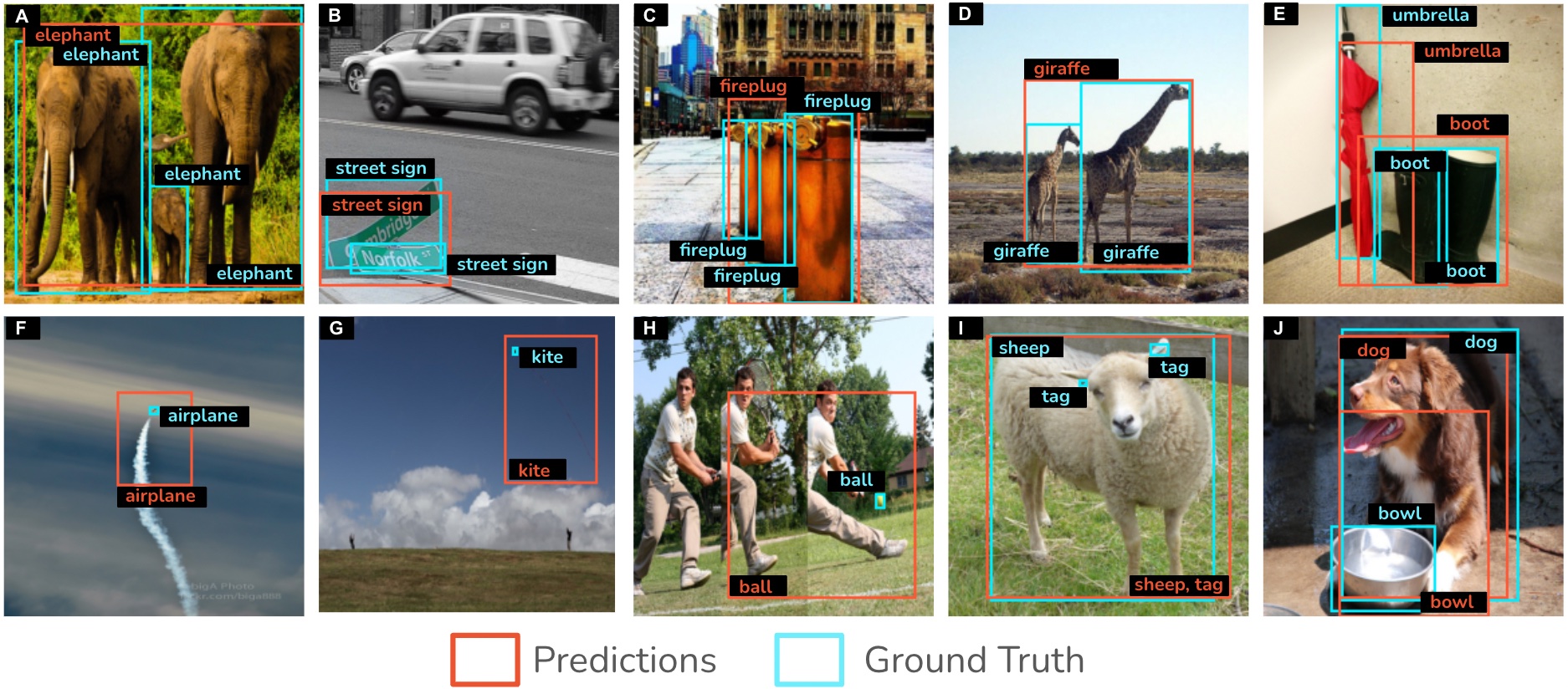}
    \caption{Typical failure cases: Due to the minimalistic design of our method, the PIN enhanced VLM cannot localise multiple instances of the same class (A-E). Often the VLM draws a bounding box around all objects of the same instance. Additionally, keeping the original input resolution of 224 from the VLM limits our ability to effectively manage very small objects (D-I).} \label{fig:failure_cases}
\end{figure*}

We visualize typical failure cases of our model in Fig.~\ref{fig:failure_cases}.
As discussed in the limitation section, our model cannot effectively localise multiple instances from the same object due to our simplistic training procedure. We found that the model typically handles those cases by drawing a bounding box around all instances from the same class which can be seen in Fig.~\ref{fig:failure_cases}A-E.
As we keep the original input resolution of the OpenFlamingo~\cite{openflamingo_awadalla} VLM of 224, our model struggles to localise these objects with a tight bounding box (Fig.~\ref{fig:failure_cases}F-I) since the object spans only across a few pixels. 

\section{Additional implementation details}
Our synthetic training and validation datasets are created from 1,116 object categories, based on LVIS, with overall 56,064 images generated by Stable Diffusion. These categories exclude those of COCO and PVOC to enable measuring truly zero-shot localisation performance. A different set of $81$ categories (which includes the COCO and PVOC classes), amounting to 4,296 images, is reserved for zero-shot evaluation. The dataset averages $50.43\pm12.11$ images per object category. 
For pasting objects onto the background images, we find dividing the images into grids of 16$\times$16 worked best for OpenFlamingo, 14$\times$14 for BLIP-2, aligning with the shapes of the vision embedding. Thus, the network only needs to predict numbers between 0 and 224 in steps of the grid size, simplifying the task at hand. This also leads to bounding boxes not being perfectly precise around the inquired object, though, it has better performance than the model trained on a grid size matching image size. For RefCOCO, we extend our synthetic dataset with positional referral expressions. For that, we increase the likelihood of sampling the same object type to 0.7. We still randomly select one of the pasted objects for training, yet, when sampling an object for which its object type occurs multiple times in the image, we add a positional referral to it. These are computed by measuring the axes with the highest difference between the center points of the objects. Then, we extend the prompt with e.g. 'left person', or 'person on left' for left, right, top, and bottom.




%% file: tables/ablation3.tex
\begin{table*}[t]
\centering
\begin{tabular}{cc  cccc }
\toprule
& $\#$ layers in $\psi$ & S embedding& mIoU & $\text{mIoU}_M$ & $\text{mIoU}_L$\\
\midrule
(a)& 1 &sinusoidal & 0.34 & 0.25 & 0.57  \\
(b)& 2 &sinusoidal &\B 0.35 & \B0.26 & \B0.59  \\
(c)& 3 &sinusoidal & 0.33 & 0.24 & 0.56  \\
\midrule
(d)& 2 & sinusoidal & 0.35 & 0.26 & 0.59  \\
(e)& 2 & learned &0.35 & 0.27 & 0.59  \\
\bottomrule
\end{tabular}
\caption{Ablation on the number of layers in $\psi$ and the type of positional embedding S used in PIN evaluated on COCO. The best performance is obtained with only 2 layers in $\psi$ and sinusoidal vs learned positional embeddings for $S$ leads to the same results. }
\label{tab:ablation3}
\end{table*}

%% file: tables/ablations1.tex
\begin{table}[t]
\centering
\resizebox{0.45\textwidth}{!}{%
\begin{tabular}{ll  cccc }
\toprule
& Background & $o_{\text{max}} $ & mIoU & $\text{mIoU}_M$ & $\text{mIoU}_L$\\
\midrule
(a)& White &0.5 & 0.24 & 0.12 & 0.48  \\
(b)& BG-20k~\cite{bg20k} &0.5  & 0.35 & 0.26 & 0.59  \\
\midrule
(c)& BG-20k~\cite{bg20k} & 0.0 & 0.33 & 0.26 & 0.56  \\
(d)& BG-20k~\cite{bg20k} & 0.5 & 0.35 & 0.26 & 0.59  \\
\bottomrule
\end{tabular}
}
\caption{Ablation on choice of background image 
and overlap between objects ($o_{\text{max}}$) on COCO. 
Realistic background images and allowing for overlap between the pasted objects improves localisation performance. }
\label{tab:ablation2}
\vspace{-0.4cm}
\end{table}